\documentclass{article}

\usepackage{arxiv}

\usepackage[utf8]{inputenc} %
\usepackage[T1]{fontenc}    %
\usepackage{hyperref}       %
\usepackage{url}            %
\usepackage{booktabs}       %
\usepackage{amsfonts}       %
\usepackage{nicefrac}       %
\usepackage{microtype}      %
\usepackage{lipsum}		%
\usepackage{graphicx}

\usepackage{xcolor}
\usepackage{amsmath}
\usepackage{float}
\usepackage{multicol}
\usepackage{multirow}
\usepackage{subcaption}
\usepackage{todonotes}
\usepackage{comment}

\newcommand{\RNum}[1]{\uppercase\expandafter{\romannumeral #1\relax}}%
\usepackage{algorithmic}
\usepackage[algoruled,boxed,lined]{algorithm2e}
\usepackage[numbers,sort]{natbib}
\usepackage{float}
\floatstyle{plaintop}
\restylefloat{table}
\usepackage[flushleft]{threeparttable}

\title{Quantifying the alignment of graph and features in deep learning}

\date{} 					%

\author{ 
	Yifan Qian \\
  School of Business and Management\\
  Queen Mary University of London\\
  London, UK \\
  \texttt{y.qian@qmul.ac.uk} \\
	\And
	Paul Expert \\
  School of Public Health\\
  Imperial College London\\
  London, UK \\
  \texttt{paul.expert08@imperial.ac.uk} \\
  \And
  Tom Rieu \\
  Facebook \\
  London, UK \\
  \texttt{rieu@fb.com} \\
  \And
  Pietro Panzarasa\thanks{Corresponding authors} \\
  School of Business and Management\\
  Queen Mary University of London\\
  London, UK \\
  \texttt{p.panzarasa@qmul.ac.uk} \\
  \And
  Mauricio Barahona\footnotemark[1] \\
  Department of Mathematics\\
  Imperial College London\\
  London, UK \\
  \texttt{m.barahona@imperial.ac.uk} \\
}

\date{}

\renewcommand{\headeright}{}
\renewcommand{\undertitle}{}

\begin{document}
\maketitle

\begin{abstract}
We show that the classification performance of graph convolutional networks (GCNs) is related to the alignment between features, graph, and ground truth, which we quantify using a subspace alignment measure (SAM) corresponding to the Frobenius norm of the matrix of pairwise chordal distances between three subspaces associated with features, graph, and ground truth. The proposed measure is based on the principal angles between subspaces and has both spectral and geometrical interpretations. We showcase the relationship between the SAM and the classification performance through the study of limiting cases of GCNs and systematic randomizations of both features and graph structure applied to a constructive example and several examples of citation networks of different origins. The analysis also reveals the relative importance of the graph and features for classification purposes.
\end{abstract}


\section{INTRODUCTION\label{sec:introduction}}
Deep learning encompasses a broad class of machine learning methods that use multiple layers of nonlinear processing units in order to learn multilevel representations for detection or classification tasks~\cite{lecun2015deep,goodfellow2016deep,schmidhuber2015deep,bronstein2017geometric,deng2014deep}. The main realizations of deep multi-layer architectures are the so-called deep neural networks (DNNs), which correspond to artificial neural networks (ANNs) with multiple layers between input and output layers. DNNs have been shown to perform successfully in processing a variety of signals with an underlying Euclidean or grid-like structure, such as speech, images and videos. Signals with an underlying Euclidean structure usually come in the form of multiple arrays~\cite{lecun2015deep} and are known for their statistical properties such as locality, stationarity and hierarchical compositionality from local statistics~\cite{simoncelli2001natural,field1989statistics}. For instance, %
an image can be seen as a function on Euclidean space (the 2-D plane) sampled from a grid. In this setting, locality is a consequence of local connections, stationarity results from shift-invariance, and compositionality stems from the intrinsic multi-resolution structure of many images~\cite{bronstein2017geometric}. 
It has been suggested that such statistical properties can be exploited by convolutional architectures via DNNs, namely (deep) convolutional neural networks (CNNs)~\cite{lecun1998gradient,lecun1990handwritten,bruna2013invariant} which are based on four main ideas: local connections, shared weights, pooling, and multiple layers~\cite{lecun2015deep}. The role of the convolutional layer in a typical CNN architecture is to detect local features from the previous layer that are shared across the image domain, thus largely reducing the parameters compared with traditional fully connected feed-forward ANNs.

Although deep learning models, and in particular CNNs, have achieved highly improved performance on data characterized by an underlying Euclidean structure, many real-world data sets do not have a natural and direct connection with a Euclidean space. Recently there has been interest in extending deep learning techniques to non-Euclidean domains, such as graphs and manifolds~\cite{bronstein2017geometric}.
An archetypal example is social networks, which can be represented as graphs with users as nodes and edges representing social ties between them. In biology, gene regulatory networks represent relationships between genes encoding proteins that can up- or down-regulate the expression of other genes. 
In this paper, we illustrate our results through examples stemming from another kind of relational data with no discernible Euclidean structure, yet with a clear graph formulation, namely citation networks, where nodes represent documents and an edge is established if one document cites the other~\cite{lazer2009life}.

To address the challenge of extending deep learning techniques to graph-structured data, a new class of deep learning algorithms, broadly named graph neural networks (GNNs), has been recently proposed~\cite{xu2018how,hamilton2017inductive,bronstein2017geometric}. In this setting, each node of the graph represents a sample, which is described by a feature vector, and we are additionally provided with relational information between the samples that can be formalized as a graph. GNNs are well suited to node (i.e., sample) classification tasks.
For a recent survey of this fast-growing field, see~\cite{wu2020comprehensive}. 

Generalizing convolutions to non-Euclidean domains is not straightforward~\cite{defferrard2016convolutional}. Recently, 
graph convolutional networks (GCNs) have been proposed~\cite{kipf2017semi} as a subclass of GNNs with convolutional properties.
The GCN architecture combines the full relational information from the graph together with the node features to accomplish the classification task, using the ground truth class assignment of a small subset of nodes during the training phase.
GCNs have shown improved performance for semi-supervised classification of documents (described by their text) into topic areas, outperforming methods that rely exclusively on text information without the use of any citation information, e.g., multilayer perceptron (MLP)~\cite{kipf2017semi}.

However, we would not expect such an improvement to be universal. In some cases, the additional information provided by the graph (i.e., the edges) might not be consistent with the similarities between the features of the nodes. In particular, in the case of citation graphs, it is not always the case that documents cite other documents that are similar in content. As we will show below with some illustrative data sets, in those cases the conflicting information provided by the graph means that a graph-less MLP approach outperforms GCN. 
Here, we explore the relative importance of the graph with respect to the features for classification purposes, and propose a geometric measure based on subspace alignment to explain the relative performance of GCN against different limiting cases.

Our hypothesis is that a degree of alignment among the three layers of information available (i.e., the features, the graph and the ground truth) is needed for GCN to perform well, and that any degradation in the information content leads to an increased misalignment of the layers and worsened performance. We will first use randomization schemes to show that the systematic degradation of the information contained in the graph and the features leads to a progressive worsening of GCN performance. Second, we propose a simple spectral alignment measure, and show that this measure correlates with the classification performance in a number of data sets: (i) a constructive example built to illustrate our work; (ii) CORA, a well-known citation network benchmark; (iii) AMiner, a newly constructed citation network data set; and (iv) two subsets of Wikipedia: Wikipedia~\RNum{1}, where GCN outperforms MLP, and Wikipedia~\RNum{2}, where instead MLP outperforms GCN.

\section{RELATED WORK\label{sec:related_work}}
\subsection{Neural Networks on Graphs}
The first attempt to generalize neural networks on graphs can be traced back to Gori~\textit{et al.}~\cite{gori2005new}, who proposed a scheme combining recurrent neural networks (RNNs) and random walk models. Their method requires the repeated application of contraction maps as propagation functions until the node representations reach a stable fixed point. This method, however, did not attract much attention when it was proposed. With the current surge of interest in deep learning, this work has been reappraised in a new and modern form:~\cite{li2015gated} introduced modern techniques for RNN training based on the original graph neural network framework, whereas~\cite{duvenaud2015convolutional} proposed a convolution-like propagation rule on graphs and methods for graph-level classification.
Non-spectral methods have also been successfully proposed. For example,~\cite{atwood2016diffusion} shows how diffusion-based representations can be learned from graph-structured data and used as the basis for node classification by introducing a diffusion-convolution operation. Niepert~\textit{et al.}~\cite{niepert2016learning} convert graphs locally into sequences fed into a conventional 1D CNN, which needs the definition of a node ordering in a pre-processing step.

The first formulation of convolutional neural networks on graphs (GCNNs) was proposed by Bruna~\textit{et al.}~\cite{bruna2014spectral}. These researchers applied the definition of convolutions to the spectral domain of the graph Laplacian. While being theoretically salient, this method is unfortunately impractical due to its computational complexity. This drawback was addressed by subsequent studies~\cite{defferrard2016convolutional}. In particular,~\cite{defferrard2016convolutional} leveraged fast localized convolutions with Chebyshev polynomials.
In~\cite{kipf2017semi}, a GCN architecture was proposed via a first-order approximation of localized spectral filters on graphs. In that work, Kipf and Welling considered the task of semi-supervised transductive node classification where labels are only available for a small number of nodes. Starting with a feature matrix $X$ and a network adjacency matrix $A$, they encoded the graph structure directly using a neural network model $f(X, A)$, and trained on a supervised target loss function $\mathcal{L}$ computed over the subset of nodes with known labels. Their proposed GCN was shown to achieve improved accuracy in classification tasks 
on several benchmark citation networks and on a knowledge graph data set. In our study, we examine how the properties of features and the graph interact in the model proposed by Kipf and Welling for semi-supervised transductive node classification in citation networks. The architecture and propagation rules of this method are detailed in Section~\ref{sec:methodGCNs}.

\subsection{Spectral Graph Convolutions}
We now present briefly the key insights introduced by Bruna~\textit{et al.}~\cite{bruna2014spectral} to extend CNNs to the non-Euclidean domain. For an extensive recent review, the reader should refer to~\cite{bronstein2017geometric}.

We study GCNs in the context of a classification task for $N$ samples. Each sample is described by a $C^{0}$-dimensional feature vector, which is conveniently arranged into the feature matrix $X \in R^{N\times C^{0}}$.
Each sample is also associated with the node of a given graph $\mathcal{G}$ with $N$ nodes, with edges representing additional relational (symmetric) information. This undirected graph is described by the adjacency matrix $A \in R^{N\times N}$. The ground truth assignment of each node to one of $F$ classes is encoded into a 0-1 membership matrix $Y \in R^{N\times F}$. 

The main hurdle is the definition of a convolution operation on a graph between a filter $g_{w}$ and the node features $X$. This can be achieved by expressing $g_{w}$ onto a basis encoding information about the graph, e.g., the adjacency matrix $A$ or the Laplacian $L=D-A$, where $D=\text{diag}(A\mathbf{1})$. This real symmetric matrix has an eigendecomposition $L=U\Lambda U^{T}$, where $U$ is the matrix of column eigenvectors with associated eigenvalues collected in the diagonal matrix $\Lambda$. The filters can then be expressed in the eigenbasis $U$ of $L$:
\begin{equation}
g_{w}=Ug_{w}(\Lambda)U^{T},
\end{equation}
with the convolution between filter and signal given by:
\begin{equation}
g_{w}\star X=Ug_{w}(\Lambda)U^{T}X.
\end{equation}
The signal is thus projected onto the space of the graph, filtered in the frequency domain, and projected back onto the~nodes.
\subsection{Graph Convolutional Networks\label{sec:methodGCNs}}
A GCN is a semisupervised method, in which a small subset of the node ground truth labels are used in the training phase to infer the class of unlabeled nodes. This type of learning paradigm, where only a small amount of labeled data is available, therefore lies between supervised and unsupervised learning.

Furthermore, the model architecture, and thus the learning, depend explicitly on the structure of the network. Hence the addition of any new data point (i.e., a new node in the network) will require a retraining of the model. GCNs are, therefore, an example of a transductive learning paradigm, where the classifier cannot be generalized to data it has not already seen.
Node classification using a GCN can be seen as a label propagation task: given a set of seed nodes with known labels, the task is to predict which label will be assigned to the unlabeled nodes given a certain topology and attributes.

\paragraph*{Layerwise Propagation Rule and Multilayer Architecture}
\label{sec:twolayerGCN}
Our study uses the multilayer GCN proposed in~\cite{kipf2017semi}.
Given the matrix $X$ with sample features and the (undirected) adjacency matrix $A$ of the graph $\mathcal{G}$ encoding relational information between the samples,
the propagation rule between layers $\ell$ and $\ell+1$ (of size $C^{\ell}$ and $C^{\ell+1}$, respectively) is given by:
\begin{equation}
H^{\ell+1} = \sigma^\ell\left(\widehat{A}H^{\ell}W^{\ell}\right),
\label{eq:layer_propagation}
\end{equation}
where 
$H^{\ell}\in R^{N\times C^{\ell}}$ and $H^{\ell+1} \in R^{N\times C^{\ell+1}}$ are matrices of activation in the $\ell^{th}$ and $(\ell+1)^{th}$ layers, respectively;  
$\sigma^\ell(\cdot)$ is the threshold activation function for layer $\ell$; and the weights connecting layers $\ell$ and $\ell+1$ are stored in the matrix $W^{\ell}\in R^{C^{\ell}\times C^{\ell+1}}$. Note that the input layer contains the feature matrix $H^{0}\equiv X$. 

The graph is encoded in
$\widehat{A}=\tilde{D}^{-1/2}\tilde{A}\tilde{D}^{-1/2}$, where $\tilde{A} = A + I_{N}$ is the adjacency matrix of a graph with added self-loops, $I_{N}$ is the identity matrix, and $\tilde{D} = \text{diag}(\tilde{A} \mathbf{1})$ is a diagonal matrix containing the degrees of $\tilde{A}$. In the remainder of this work (and to ensure comparability with the results in~\cite{kipf2017semi}), we use $\widehat{A}$ as the descriptor of the graph $\mathcal{G}$. 

\begin{figure}[htbp!]
\centering
\includegraphics[width=0.5\textwidth]{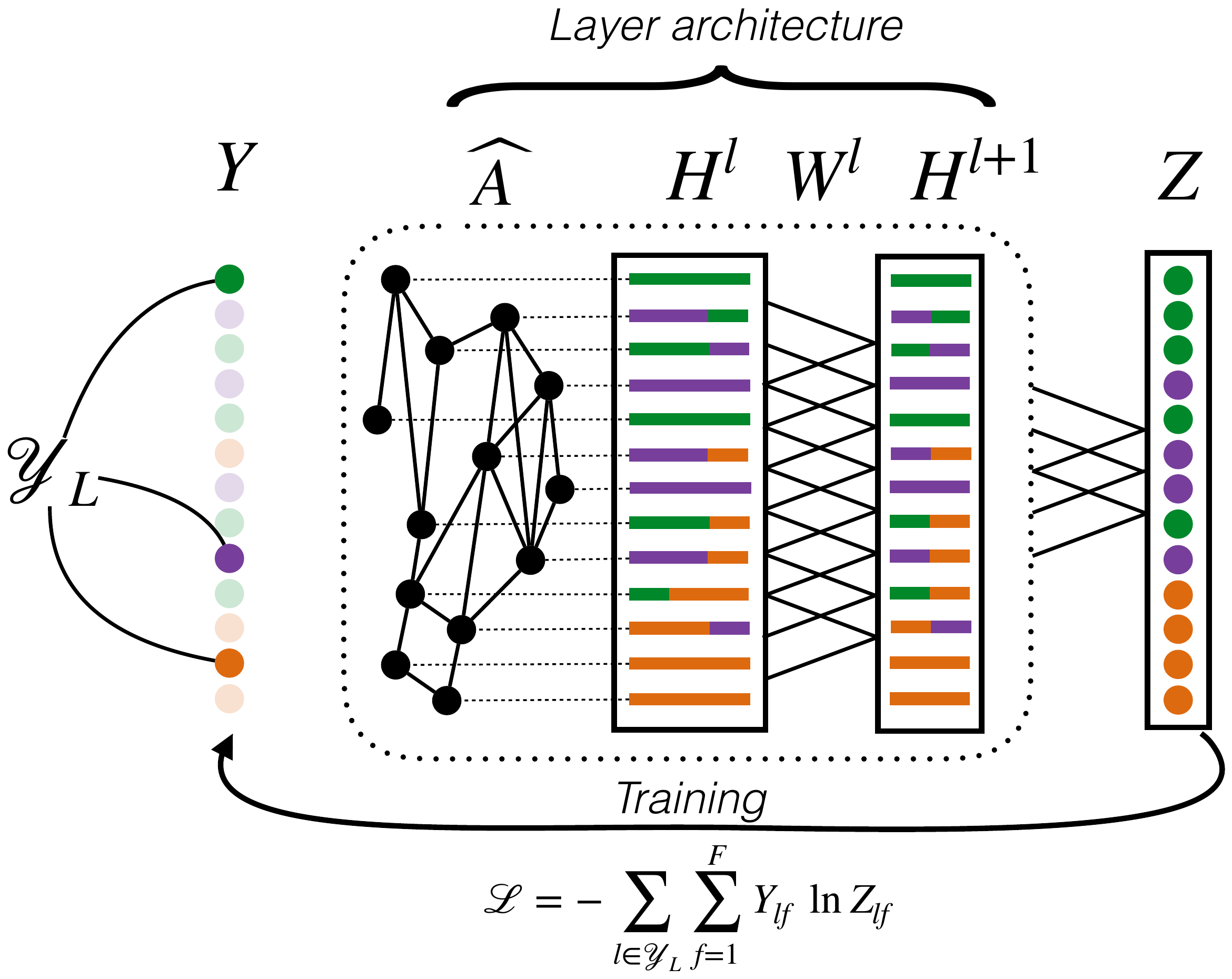}
\caption{\textbf{Schematic of GCN used.} The graph $\widehat{A}$ is applied to the input of each layer $\ell$ before it is funneled into the input of layer $\ell+1$. The process is repeated until the output has dimension $N\times F$ and produces a predicted class assignment. During the training phase, the predicted assignments are compared against a subset of values $\mathcal{Y}_L$ of the ground truth.}
\label{fig:gcn_architecture}
\end{figure}

Following \cite{kipf2017semi}, 
we implement a two-layer GCN with propagation rule~\eqref{eq:layer_propagation} and different activation functions for each layer, i.e., a rectified linear unit (ReLU) for the first layer and a softmax unit for the output layer:
\begin{align}
\sigma^0 &:\mathrm{ReLU}(x_{i})=\max(x_{i},0)  \\
\sigma^1 &:\mathrm{softmax}(x)_{i} = \frac{\exp(x_{i})}{\sum_{i} \exp(x_{i})}, 
\end{align}
where $x$ is a vector.
The model then takes the simple form:
\begin{equation}
Z = f(X,A) %
 = \mathrm{softmax}(\widehat{A}~\mathrm{ReLU}(\widehat{A}XW^{0})~W^{1}),\\
\label{eq:two_layers_rule}
\end{equation}
where the softmax function is applied row-wise and the ReLU is applied element-wise. 
Note there is only one hidden layer with $C^1$ units. Hence 
$W^{0} \in R^{C^{0}\times C^{1}}$ maps the input with $C^0$ features to the hidden layer 
and $W^{1} \in R^{C^{1}\times C^{2}}$ maps these hidden units to the output layer with $C^{2}=F$ units, corresponding to the number of classes of the ground truth. 
In this semi-supervised multiclass classification, the cross-entropy error over all labeled instances is evaluated as follows:
\begin{equation}
        \mathcal{L} = -\displaystyle\sum_{l\in \mathcal{Y}_{L}}\displaystyle\sum_{f=1}^{F} Y_{lf} \, \ln{Z_{lf}},
    \label{eq:cross_entropy_error}
\end{equation}
where $\mathcal{Y}_{L}$ is the set of nodes that have labels. The 
weights of the neural network ($W^{0}$ and $W^{1}$) are trained using gradient descent to minimize the loss~$\mathcal{L}$.
A visual summary of the GCN architecture is shown in Fig.~\ref{fig:gcn_architecture}. The reader is referred to~\cite{kipf2017semi} for details and in-depth analysis. Although GCN was introduced as a simplified form of spectral-based GNNs, it shows a natural connection with spatial-based GNNs, in which graph convolutions are defined by information propagation. Hence our study of the alignment of graph and features is related more widely to graph-feature correlations, such as spatial autocorrelations measured with, e,g,, Moran's and Geary's indices~\cite{de1984extreme,waldhor06moran} that capture how the features of nodes influence each other via network structure.

\section{METHODS\label{sec:methods}}
\subsection{Randomization Strategies\label{sec:randomization_strategy}}
To test the hypothesis that a degree of alignment across information layers is crucial for a good classification performance of GCN, we gradually randomize the node features, the node connectivity, or both. For the randomization to give a meaningful notion of alignment, at least one ingredient needs to be kept constant. Since we focus on the alignment of graph and features, we keep the ground truth constant.

\subsubsection{Randomization of the Graph}
The edges of the graph are randomized by rewiring a percentage $p_{\widehat{A}}$ %
of edge stubs (i.e., ``half-edges'') under the constraint that the degree distribution remains unchanged. This randomization strategy is described in Algorithm~\ref{algo:random_graph} which is based on the configuration model~\cite{newman2003structure}. Once a randomized realization of the graph is produced, the corresponding $\widehat{A}$ is computed.

\begin{algorithm}[htbp!]
\KwIn{A graph $G(V,E)$, where $V$ is the set of nodes and $E$ is the set of edges, and a randomization percentage $0\leq p_{\widehat{A}} \leq 100$.}
\KwOut{A randomized graph $G_{p_{\widehat{A}}}(V,E')$}
\BlankLine
1.Choose a random subset of edges $E_{r}$ from $E$ with $|E_{r}|=\left\lfloor|E|\times p_{\widehat{A}} /100 \right \rfloor$, and denote the unrandomized edges in $E$ as $E_{u}$.
\BlankLine
2. Obtain the degree sequence of nodes from $E_{r}$, and build a stub list $l_{s}$ based on the degree sequence.
\BlankLine
3. Obtain a randomized stub list $l'_{s}$ by shuffling $l_{s}$, and randomized edges $E'_{r}$ by connecting the stubs in the corresponding positions of the two stub lists~$l_{s}$~and~$l'_{s}$.
\BlankLine
4. Compute $E_{u} \cup E'_{r}$, remove multiedges and self-loops, and obtain the final edge set E'.
\BlankLine
5. Generate randomized graph $G_{p_{\widehat{A}}}(V,E')$ from node set V and edge set $E'$.
\caption{Randomization of the Graph}
\label{algo:random_graph}
\end{algorithm}

\subsubsection{Randomization of the Features}
The features were randomized by swapping feature vectors between a percentage $p_{X}$ %
of randomly chosen nodes following the procedure described in Algorithm~\ref{algo:random_features}. 
\begin{algorithm}[htbp!]
\KwIn{A feature matrix $X \in R^{N\times C^{0}}$, and a randomization percentage $0\leq p_X\leq 100$.}
\KwOut{A randomized feature matrix $X_{p_{X}}\in R^{N\times C^{0}}$}
\BlankLine
1. Choose at random $N_{r}$ rows from $X$, where 
$N_{r}=\left\lfloor N \, p_X/100 \right\rfloor$.
\BlankLine
2. Swap randomly the $N_{r}$ rows to obtain $X_{p_{X}}$.
\caption{Randomization of the Features}
\label{algo:random_features}
\end{algorithm}

A fundamental difference between the two randomization schemes is that the graph randomization alters its spectral properties as it gradually destroys the graph structure, whereas the randomization of the features preserves its spectral properties in the principal component analysis (PCA) sense, i.e., the principal values are the same but the loadings on the components are swapped. Hence the feature randomization still alters the classification performance because the features are re-assigned to nodes that have a different environment, thereby changing the result of the convolution operation defined by the $H^{\ell}$ activation matrices~\eqref{eq:layer_propagation}.  

\subsection{Limiting Cases\label{sec:limit}}
To interrogate the role that the graph plays in the classification performance of a GCN, it is instructive to consider three limiting cases:
\begin{itemize}
\item \textit{No Graph:} $A=\mathbf{0}\mathbf{0}^{T}$. If we remove all the edges in the graph, the classifier becomes equivalent to an MLP, a classic feed-forward ANN. The classification is based solely on the information contained in the features, as no graph structure is present to guide the label propagation.

\item  \textit{Complete Graph:} $A=\mathbf{1}\mathbf{1}^{T} - I_{N}$. 
In this case, the mixing of features is immediate and homogeneous, corresponding to a mean field approximation of the information contained in the features.

\item \textit{No Features:} $X=I_{N}$. In this case, the label propagation and assignment are purely based on graph topology.  
\end{itemize}
An illustration of these limiting cases can be found in the top row of Table~\ref{table:results_gcn_and_limitCases}.

\subsection{Spectral Alignment Measure}
In order to quantify the alignment between the features, the graph and the ground truth, we propose a measure based on the chordal distance between subspaces, as follows. 

\subsubsection{Chordal Distance Between Two Subspaces}
Recent work by Ye and Lim~\cite{ye2016schubert} has shown that the distance between two subspaces of different dimension in $\mathbb{R}^{n}$ is necessarily defined in terms of their principal angles. 

Let $\mathcal{A}$ and $\mathcal{B}$ be two subspaces of the ambient space $\mathbb{R}^{n}$
with dimensions $\alpha$ and $\beta$, respectively, with $\alpha \leq \beta < n$. The principal angles between $\mathcal{A}$ and $\mathcal{B}$ denoted $0 \leq \theta_{1}\leq\theta_{2}\leq...\leq\theta_{\alpha} \leq \frac{\pi}{2}$ are defined recursively as follows~\cite{bjorck1973numerical,golub2012matrix}:
\begin{align*}
\theta_{1}&=\min_{a_{1}\in \mathcal{A}, b_{1}\in \mathcal{B}} \arccos\left(\frac{|a_{1}^{T}b_{1}|}{\|a_{1}\|\|b_{1}\|}\right), &\\
\theta_{j}&=\min_{\substack{a_{j}\in \mathcal{A}, b_{j}\in \mathcal{B}\\
                          a_{j}\bot a_{1},...,a_{j-1}\\
                          b_{j}\bot b_{1},...,b_{j-1}}}
\arccos\left(\frac{|a_{j}^{T}b_{j}|}{\|a_{j}\|\|b_{j}\|}\right), \enskip j=2,...,\alpha,
\end{align*}
If the \textit{minimal} principal angle is small, then the two subspaces are nearly linearly dependent, i.e., almost perfectly aligned.
A numerically stable algorithm that computes the canonical correlations, (i.e., the cosine of the principal angles) between subspaces is given in Algorithm~\ref{algo:principal_angles}. %

\begin{algorithm}[htbp!]
\KwIn{matrices $A_{n \times \alpha}$ and $B_{n \times \beta}$ with $\alpha \leq \beta < n$.}
\KwOut{cosines of the principal angles $\theta_{1}\leq \theta_{2}\leq...\leq \theta_{\alpha}$ between $\mathcal{R}(A)$ and $\mathcal{R}(B)$, the column spaces of $A$ and $B$.}
\BlankLine
1. Find orthonormal bases $\mathcal{Q}_{A}$ and $\mathcal{Q}_{B}$ for $A$ and $B$ using the QR decomposition:
    $\mathcal{Q}_{A}^{T}\mathcal{Q}_{A} = \mathcal{Q}_{B}^{T}\mathcal{Q}_{B} = I$; $\mathcal{R}(\mathcal{Q}_{A}) = \mathcal{R}(A)$, $\mathcal{R}(\mathcal{Q}_{B}) = \mathcal{R}(B)$.
\BlankLine
2. Compute the singular value decomposition (SVD): %
$\mathcal{Q}_{A}^{T}\mathcal{Q}_{B}$ = $UCV^{T}$.
\BlankLine
3. Extract the diagonal elements of $C$: $C_{ii}=\cos \theta_i$, to obtain the canonical correlations $\{\cos\theta_{1},...,\cos\theta_{\alpha}\}$.
\caption{Principal angles~\cite{bjorck1973numerical,golub2012matrix}}
\label{algo:principal_angles}
\end{algorithm}

The principal angles are the basic ingredient of a number of well defined Grassmannian distances between subspaces~\cite{ye2016schubert}. Here we use the chordal distance
given by:%
\begin{equation}
d(\mathcal{A,B}) = \sqrt{\displaystyle\sum_{j=1}^{\alpha}{\sin^{2}\theta_{j}}}.
\label{eq:pairwise_alignment}
\end{equation}
The larger the chordal distance $d(\mathcal{A,B})$ is, the worse the alignment between the subspaces $\mathcal{A}$ and $\mathcal{B}$. %

We remark that the last inequality in $\alpha \leq \beta < n$ is strict. If a subspace spans the whole ambient space (i.e., $\beta=n$), then its distance to all other strict subspaces of $\mathbb{R}^n$ is trivially zero, as it is always possible to find a rotation that aligns the strict subspace with the whole space.

\subsubsection{Alignment Metric\label{sec:Frobenius_norm}}
Our task involves establishing the alignment between \textit{three} subspaces associated with the features $X$, the graph $\widehat{A}$, and the ground truth $Y$. 
To do so, we consider the distance matrix containing all the pairwise chordal distances: 
\begin{equation}
D(X,\widehat{A},Y) = \begin{bmatrix}
0 & d(X,\widehat{A}) & d(X,Y) \\[0.3em]
d(X,\widehat{A}) & 0 & d(\widehat{A},Y) \\[0.3em]
d(X,Y) & d(\widehat{A},Y) & 0
\end{bmatrix},
\label{eq:matrix_distance}
\end{equation}
and we take the Frobenius norm~\cite{golub2012matrix} of this matrix $D$ as our \textit{subspace alignment measure} (SAM):
\begin{equation}
\mathcal{S}(X,\widehat{A},Y) = \|D(X,\widehat{A},Y) \|_{\text{F}} = \sqrt{\sum_{i=1}^{3}\sum_{j=1}^{3}D_{ij} ^{2}}.
\label{eq:norm_matrix_distance}
\end{equation}
The larger $\|D\|_{\text{F}}$ is, the worse the alignment between the three subspaces. 
This alignment measure has a geometric interpretation related to the area of the triangle with sides $d(X,\widehat{A}), d(X,Y), d(\widehat{A},Y)$ (the smaller blue shaded triangle in Fig.~\ref{fig:pyramid}). 

\subsubsection{Determining the Dimension of the Subspaces\label{sec:generation_subspaces}}

The feature, graph and ground truth matrices $(X,\widehat{A},Y)$ are associated with subspaces of the ambient space $\mathbb{R}^N$, where $N$ is the number of nodes (or samples). These subspaces are spanned by: the eigenvectors of $\widehat{A}$, the principal components of the feature matrix $X$, and the principal components of the ground truth matrix $Y$, respectively~\cite{von2007tutorial}. 
The dimension of the graph subspace is $N$; the dimension of the feature subspace is the number of features $C^{0}<N$ (in our examples); and the dimension of the ground truth subspace is the number of classes~$F<C^{0}<N$. 

\begin{figure}[htbp!]
\centering
\includegraphics[width=0.5\textwidth]{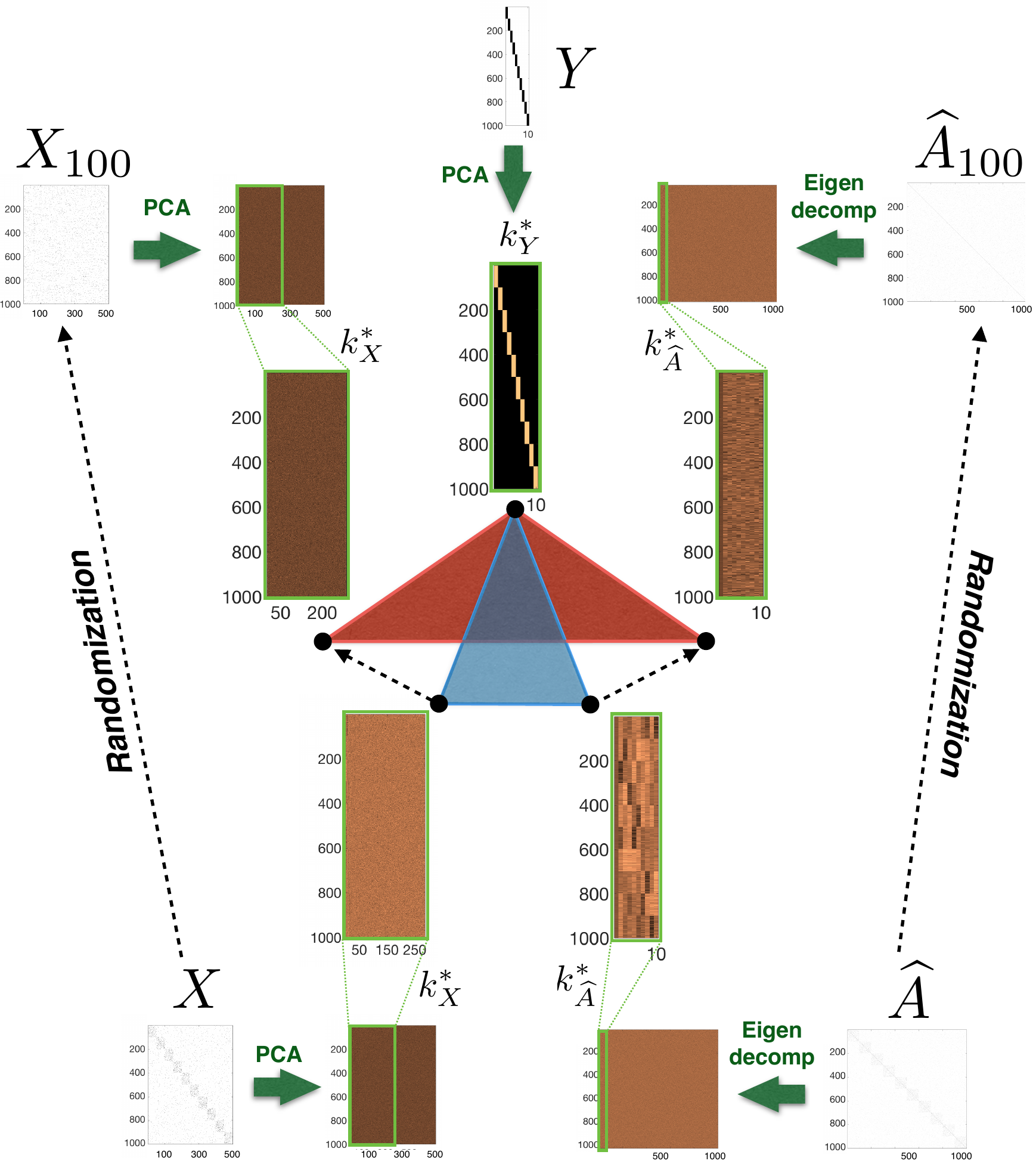}
\caption{\textbf{Method to determine relevant subspaces~\eqref{eq:dimension_subspaces}}. Using the constructive example, we illustrate the subspaces representing features, graph and ground truth. The feature and ground truth matrices are decomposed via PCA %
and the graph matrix is similarly eigendecomposed.
Fixing $k_{Y}^{*}=F$, we optimize~\eqref{eq:dimension_subspaces} to find the dimensions $k_{X}^{*}$ and $k_{\widehat{A}}^{*}$ that maximize the difference between the area of the blue triangle, which reflects the alignment of the three subspaces $(X,\widehat{A},Y)$ of the original data, and the area of the red triangle, which corresponds to the alignment of the subspaces $(X_{100},\widehat{A}_{100},Y)$ of the fully randomized data. 
The edges of the triangles correspond to the pairwise chordal distances (e.g., the base of the blue triangle corresponds to $d(X,\widehat{A})$).
}
\label{fig:pyramid}
\end{figure}

The pairwise chordal distances $D_{ij}$ in~\eqref{eq:matrix_distance} are computed from a number of minimal angles, corresponding to the smaller of the two dimensions of the subspaces being compared. 
Hence the dimensions of the subspaces $(k_X, k_{\widehat{A}}, k_Y)$ need to be defined to compute the distance matrix $D$.  
Here, we are interested in finding low dimensional subspaces of features, graph and ground truth with dimensions $(k^*_X, k^*_{\widehat{A}}, k^*_Y)$ such that they provide maximum discriminatory power between the original problem and the fully randomized (null) model.
To do this, we propose the following criterion:
\begin{align}
    k^{*}_{Y}&=F    \label{eq:dimension_subspaces} \\
  (k_{X}^{*},k_{\widehat{A}}^{*})&=
  \underset{k_{X},k_{\widehat{A}}}{\max}
  \left(\|D(X_{100},\widehat{A}_{100},Y)\|_{\text{F}}-
  \|D(X,\widehat{A},Y)\|_{\text{F}}\right). \nonumber
\end{align}
We choose $k^{*}_{Y}$ equal to the number of ground truth classes since they are non-overlapping~\cite{von2007tutorial}.
Our optimization selects $k_{X}^{*}$ and $k_{\widehat{A}}^{*}$ such that the difference in alignment between the original problem with no randomization ($p_X=p_{\widehat{A}}=0$)
and an ensemble of 100 fully randomized (feature and graph, $p_X=p_{\widehat{A}}=100$) problems is maximized (see SI for details on the optimization scheme). 
This criterion maximizes the range of values that $||D||_{\text{F}}$ can take, thus augmenting the discriminatory power of the alignment measure when finding the alignment between \textit{both} data sources and the ground truth, beyond what is expected purely at random. 
Importantly, the reduced dimension of features and graph are found simultaneously, since our objective is to quantify the alignment (or amount of shared information) contained in the three subspaces. Our criterion effectively amounts to finding the dimensions of the subspaces that maximize a difference in the surfaces of the larger (red) and smaller (blue) shaded triangles in Fig.~\ref{fig:pyramid}. 

We provide the code to compute our proposed alignment measure at \url{https://github.com/haczqyf/gcn-data-alignment}.

\section{EXPERIMENTS\label{sec:experiments}}
\subsection{Data Sets\label{sec:dataset}}
Relevant statistics of the data sets, including number of nodes and edges, dimension of feature vectors, and number of ground truth classes, are reported in Table~\ref{table:dataset_statistics}.
\begin{table}[htbp!]
\centering
\resizebox{0.5\textwidth}{!}{
\begin{tabular}{ ccccccc }
\specialrule{.1em}{.05em}{.05em}
\textbf{Data sets} & \textbf{Nodes ($N$)} & \textbf{Edges} & \textbf{Features ($C^0$)} & \textbf{Classes ($F$)}\\
\hline
Constructive & $1,000$ & $6,541$ & $500$ & $10$\\
CORA & $2,485$ & $5,069$ & $1,433$ & $7$\\
AMiner & $2,072$ & $4,299$ & $500$ & $7$\\
Wikipedia & $20,525$ & $215,056$ & $100$ & $12$\\
Wikipedia~\RNum{1} & $2,414$ & $8,163$ & $100$ & $5$\\
Wikipedia~\RNum{2} & $1,858$ & $8,444$ & $100$ & $5$\\
\specialrule{.1em}{.05em}{.05em}
\end{tabular}}
\caption{\textbf{Some statistics of the data sets in our study.}}
\label{table:dataset_statistics}
\end{table}

\subsubsection{Constructive Example}\label{sec:constructive_example}
To illustrate the alignment measure in a controlled setting, we build a constructive example, consisting of $1,000$ nodes assigned to $10$ planted communities $C_1,...,C_{10}$ of equal size. 
We then generate both a feature matrix and a graph matrix whose structures are aligned with the ground truth assignment matrix. The graph structure is generated using a stochastic block model that reproduces the ground truth structure with some noise: two nodes are connected with a probability $p_{in}=0.07$ if they belong to the same community $C_i$ and $p_{out}=0.007$ otherwise. The feature matrix is constructed in a similar way. The feature vectors are $500$ dimensional and binary, i.e., a node either possesses a feature or it does not. Each ground truth cluster is associated with $50$ features that are present with a probability of $p_{in}=0.07$. Each node also has a probability $p_{out}=0.007$ of possessing each feature characterizing other clusters. Using the same stochastic block structure for both features and graph ensures that they are maximally aligned with the ground truth. This constructive example is then randomized in a controlled way to detect the loss of alignment and the impact this loss of alignment has on the classification performance.  

\subsubsection{CORA}
The CORA data set is a benchmark for classification algorithms using text and citation data\footnote{https://linqs.soe.ucsc.edu/data}. 
Each paper is labeled as belonging to one of $7$ categories (Case\_Based, Genetic\_Algorithms, Neural\_Networks, Probabilistic\_Methods, Reinforcement\_Learning, Rule\_Learning, and Theory), which gives the ground truth $Y$.
The text of each paper is described by a $0/1$ vector indicating the absence/presence of words in a dictionary of $1,433$ unique words, the dimension of the feature space.
The feature matrix $X$ is made from these word vectors.
We extracted the largest connected component of this citation graph (undirected) to form the graph adjacency matrix~$A$.

\subsubsection{AMiner}
For additional comparisons, we produced a new data set with similar characteristics to CORA from the academic citation site AMiner.
AMiner is a popular scholarly social network service for research purposes only~\cite{tang2008arnetminer},
which provides an open database\footnote{https://aminer.org/data} with more than $10$ data sets encompassing researchers, conferences, and publication data.
Among these, the academic social network\footnote{https://aminer.org/aminernetwork} is the largest one and includes information on papers, citations, authors, and scientific collaborations.
In 2012 the Chinese Computer Federation (CCF) released a catalog including $10$ subfields of computer science. Using the AMiner academic social network, Qian~\textit{et al.}~\cite{qian2017citation} extracted $102,887$ papers published from 2010 to 2012, 
and mapped each paper with a unique subfield of computer science according to the publication venue.
Here, we use these assigned categories as the ground truth for a classification task. 
Using all the papers in~\cite{qian2017citation} that have both abstract and references, we created a data set of similar size to CORA. We extracted the largest connected component from the citation network of all papers in $7$ subfields (Computer systems/high performance computing, Computer networks, Network/information security, Software engineering/software/programming language, Databases/data mining/information retrieval, Theoretical computer science, and Computer graphics/multimedia) from 2010 to 2011. The resulting AMiner citation network consists of $2,072$ papers with $4,299$ edges. Just as with CORA, we treat the citations as undirected edges, and obtain an adjacency matrix $A$. We further extracted the most frequent $500$ stemmed terms from the corpus of abstracts of papers and constructed the feature matrix $X$ for AMiner using bag-of-words.

\subsubsection{Wikipedia}
As a contrasting example, we produced three data sets from the English Wikipedia.
The Wikipedia provides an interlinked corpus of documents (articles) in different fields, which `cite' each other via hyperlinks. We first constructed a large corpus of articles, consisting of a mixture of popular and random pages so as to obtain a balanced data set. We retrieved the $5,000$ most accessed articles during the week before the construction of the data set (July 2017), and an additional $20,000$ documents at random using the Wikipedia built-in random function\footnote{https://en.wikipedia.org/wiki/Wikipedia:Random}. 
The text and subcategories of each document, together with the names of documents connected to it, were obtained using the Python library~\textit{Wikipedia}\footnote{https://github.com/goldsmith/Wikipedia}. 
A few documents (e.g., those with no subcategories) 
were filtered out during this process. 
We constructed the citation network of the documents retrieved and extracted the largest connected component. The resulting citation network contained $20,525$ nodes and $215,056$ edges. The text content of each document was converted into a bag-of-words representation based on the $100$ most frequent words. To establish the ground truth, 
we used $12$ categories from the application programming interface (API) (People, Geography, Culture, Society, History, Nature, Sports, Technology, Health, Religion, Mathematics, Philosophy) and assigned each document to one of them.
As part of our investigation, we split this large Wikipedia data set into two smaller subsets of non-overlapping categories: Wikipedia~\RNum{1}, consisting of Health, Mathematics, Nature, Sports, and Technology; and Wikipedia~\RNum{2}, with the categories Culture, Geography, History, Society, and People.

All six data sets used here can be found at \url{https://github.com/haczqyf/gcn-data-alignment/tree/master/alignment/data}.

\subsection{GCN Architecture, Hyperparameters and Implementation}\label{sec:experiment_setup}
We used the GCN architecture~\cite{kipf2017semi} and implementation\footnote{https://github.com/tkipf/gcn} provided by Kipf and Welling~\cite{kipf2017semi}, and followed closely their experimental setup to train and test the GCN on our data sets. 
We used a two-layer GCN as described in Section~\ref{sec:methodGCNs} with the maximum number of training iterations (epochs) set to $400$~\cite{kingma2014adam}, a learning rate of $0.01$, and early stopping with a window size of $100$, i.e., training stops if the validation loss does not decrease for $100$ consecutive epochs.
Other hyperparameters used were: 1) dropout rate: $0.5$; 2) L2 regularization: $5 \times 10^{-4}$; and 3) number of hidden units: $16$. We initialized the weights as described in~\cite{glorot2010understanding}, and accordingly row-normalized the input feature vectors.
For the training, validation and test of the GCN, we used the following split: 1) $5$\% of instances as training set; 2) $10$\% as validation set; and 3) the remaining $85$\% as test set. We used this split for all data sets with exception of the full Wikipedia data set, where we used: 1) $3.5$\% of instances as training set; 2) $11.5$\% as validation set; and 3) the remaining $85$\% as test set. This modification of the split was necessary to ensure the instances in the training set were evenly distributed across categories.

\section{RESULTS\label{sec:results}}
The GCN performance is evaluated using the standard \textit{classification accuracy} defined as the proportion of nodes correctly classified in the test set.  

\subsection{GCN: Original Graph Versus Limiting Cases\label{sec:results_gcn_limitCases}}
For each data set in Table~\ref{table:dataset_statistics}, we trained and tested a GCN with the original graph and features matrices, and GCN models under the three limiting cases described in Section~\ref{sec:limit}. 
We computed the average accuracy of $100$ runs with random weight initializations (Table~\ref{table:results_gcn_and_limitCases}).
\begin{table}[htbp!]
\centering
\resizebox{0.75\textwidth}{!}{
\begin{threeparttable}
\begin{tabular}{lcccc}
\specialrule{.1em}{.05em}{.05em}
& \textbf{GCN (original)} & \multicolumn{3}{c}{\textbf{GCN (limiting cases)}}\\
\cmidrule(lr){2-2}\cmidrule(l){3-5}
& & No graph = MLP & No features & Complete graph\\
& & \textit{(Only features)} & \textit{(Only graph)}  & 
\textit{(Mean field)}\\
& & $A=\mathbf{0}\mathbf{0}^{T}$ &  $X=I_N$ & $A=\mathbf{1}\mathbf{1}^{T} - I_{N}$ \\
 & \raisebox{-\totalheight}{\includegraphics[width=0.11\textwidth,height=0.12\textwidth]{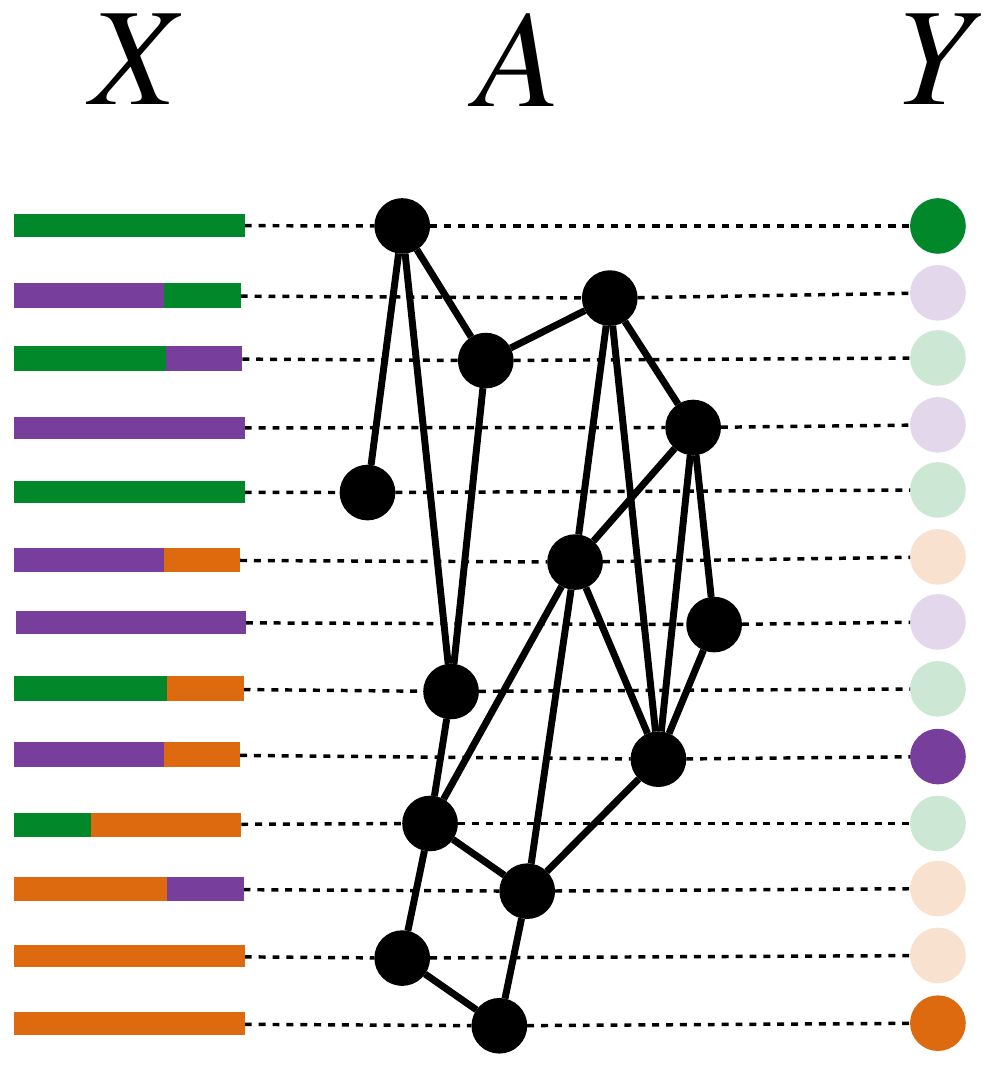}} & \raisebox{-\totalheight}{\includegraphics[width=0.11\textwidth,height=0.12\textwidth]{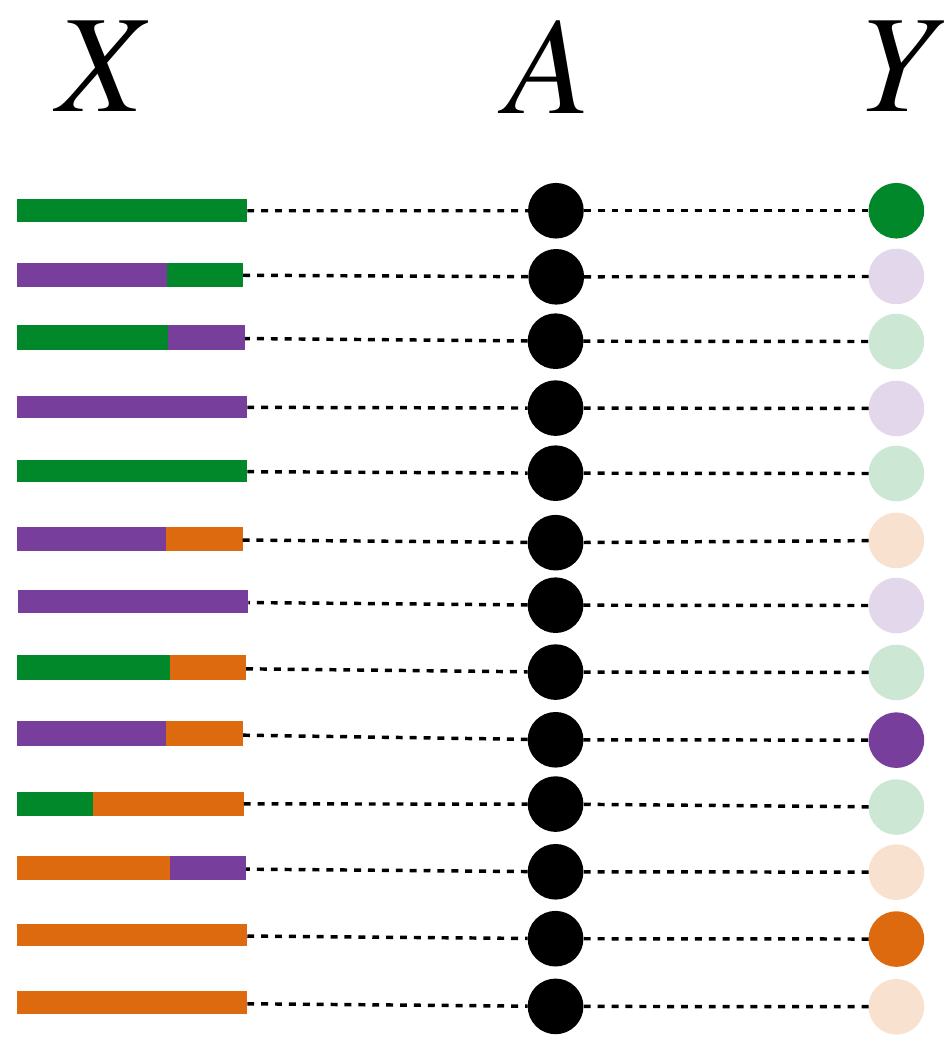}} & \raisebox{-\totalheight}{\includegraphics[width=0.11\textwidth,height=0.12\textwidth]{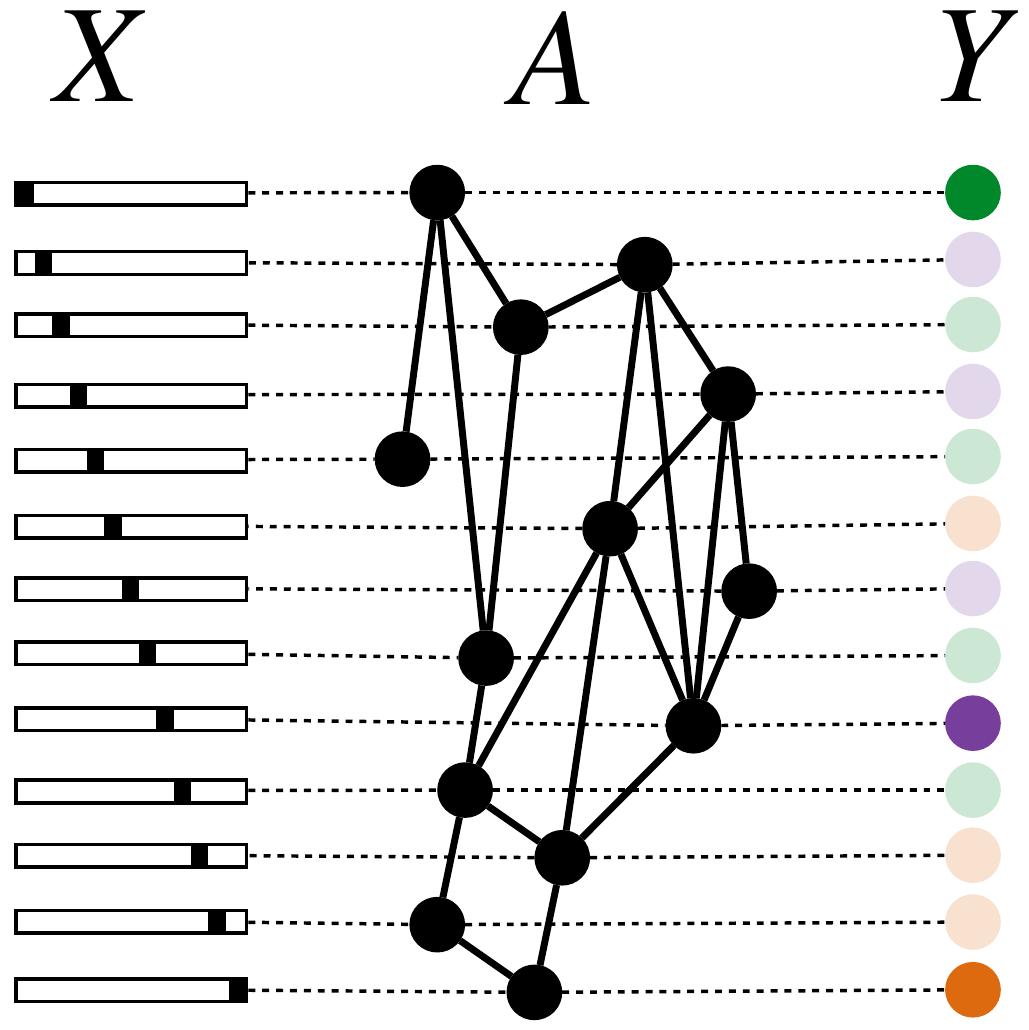}} & \raisebox{-\totalheight}{\includegraphics[width=0.11\textwidth,height=0.12\textwidth]{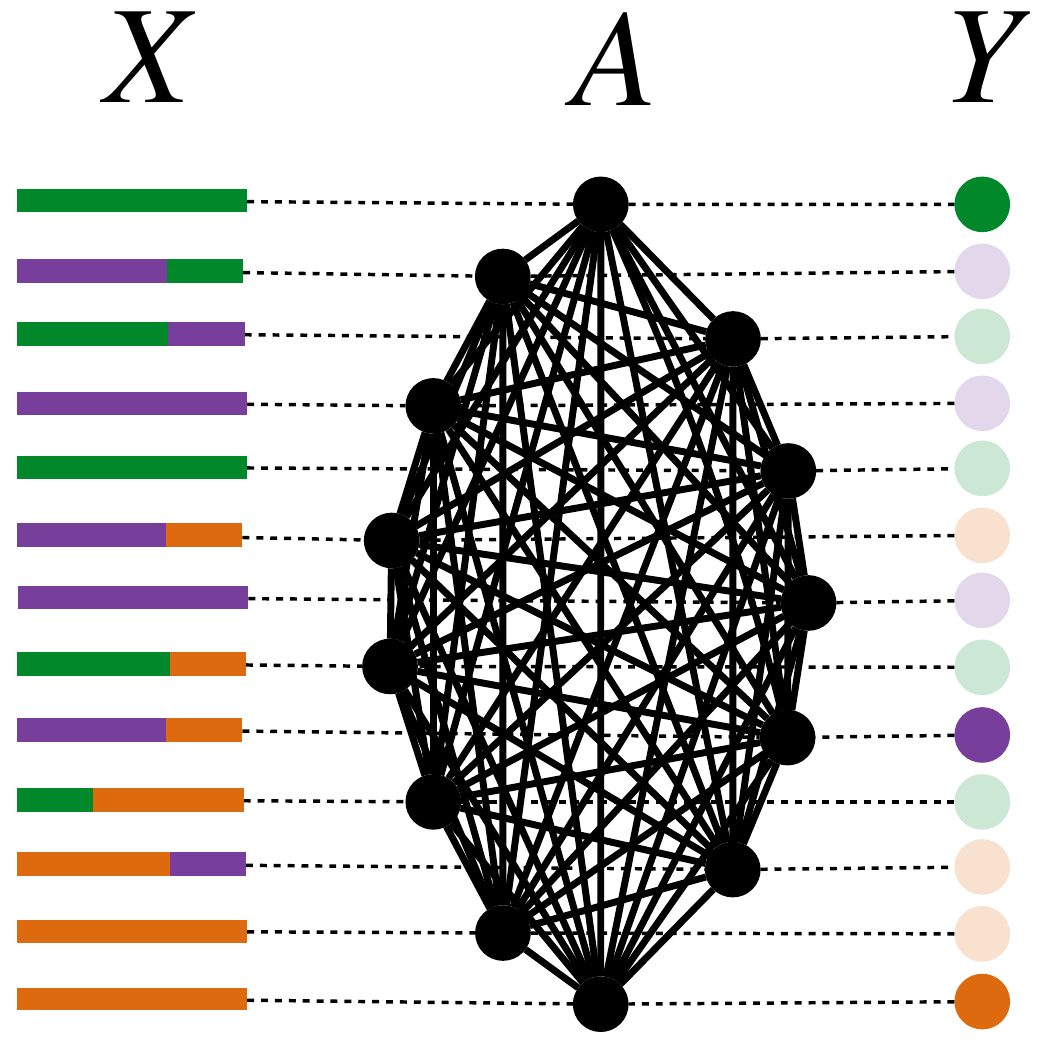}}\\
 \textbf{Data sets} & & & & \\ %
\hline
Constructive & \textbf{0.932 $\pm$ 0.006} & 0.416 $\pm$ 0.010 & 0.764 $\pm$ 0.009 & 0.100 $\pm$ 0.003 \\
CORA                 & \textbf{0.811 $\pm$ 0.005} & 0.548 $\pm$ 0.014 & 0.691 $\pm$ 0.006 & 0.121 $\pm$ 0.066 \\
AMiner               & \textbf{0.748 $\pm$ 0.005} & 0.547 $\pm$ 0.013 & 0.591 $\pm$ 0.006 & 0.123 $\pm$ 0.045 \\
Wikipedia            & 0.392 $\pm$ 0.010          & \textbf{0.450 $\pm$ 0.007} & 0.254 $\pm$ 0.037 & O.O.M.            \\
Wikipedia~\RNum{1}   & \textbf{0.861 $\pm$ 0.006} & 0.796 $\pm$ 0.005 & 0.824 $\pm$ 0.003 & 0.163 $\pm$ 0.135 \\
Wikipedia~\RNum{2}   & 0.566 $\pm$ 0.021          & \textbf{0.659 $\pm$ 0.011} & 0.347 $\pm$ 0.012 & 0.155 $\pm$ 0.176 \\
\specialrule{.1em}{.05em}{.05em}
\end{tabular}
\end{threeparttable}
}
\caption{\textbf{Classification accuracy of GCN and limiting cases for our data sets.}
The best performance is indicated in bold. Error bars are evaluated over $100$ runs. The GCN with original data performs best in most cases, but is outperformed by MLP in the full Wikipedia data set and its subset Wikipedia II.
}
\label{table:results_gcn_and_limitCases}
\end{table}

The GCN using all the information available in the features and the graph outperforms MLP (the no graph limit) except in the case of the large Wikipedia set. 
Hence using the additional information contained in the graph does not necessarily increase the performance of GCN. 
To investigate this issue further, 
we split the Wikipedia data set into two subsets: Wikipedia~\RNum{1}, with articles in topics that tend to be more self-referential (e.g., Mathematics or Technology) and Wikipedia~\RNum{2}, containing pages in areas that are less self-contained (e.g., Culture or Society). We observed that GCN outperforms MLP for Wikipedia~\RNum{1} but the opposite is still true for Wikipedia~\RNum{2}.
Finally, we also observe that the performance of ``No features'' is always lower than the performance of GCN, and, as expected, the performance of ``Complete graph'' (i.e., mean field) is very low and close to pure chance (i.e., $\sim 1/F$).

\subsection{Performance of GCN Under Randomization\label{sec:results_randomization}}
\begin{figure}[htbp!]
\centering
\includegraphics[width=0.75\textwidth]{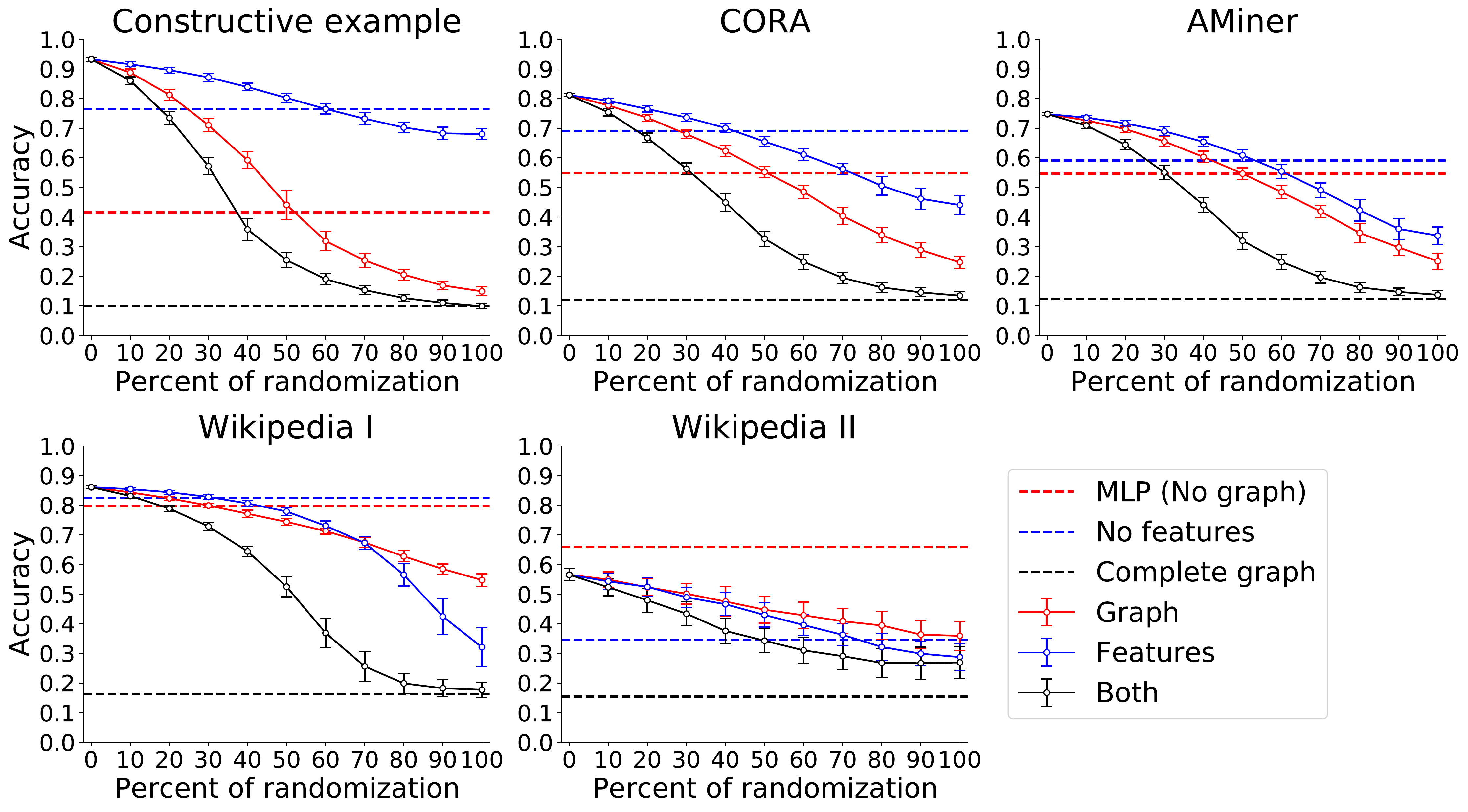}
\caption{\textbf{Degradation of classification performance as a function of randomization.} Each panel shows the degradation of the classification accuracy as a function of the randomization of graph, features and both, for a different data set. Error bars are evaluated over $100$ realizations:
for zero percent randomization, we report $100$ runs with random weight initializations; for the rest, we report $1$ run with random weight initializations for $100$ random realizations.
The horizontal lines correspond to the limiting cases in Table~\ref{table:results_gcn_and_limitCases}. 
The full Wikipedia data set was not analyzed here
since the eigendecomposition of $\widehat{A}$ needed to obtain $k^{*}_{X}, k^{*}_{\widehat{A}}$ is computationally intensive.
}
\label{fig:summary_randomization}
\end{figure}

The results above 
lead us to pose the hypothesis that a degree of synergy between features, graph and ground truth is needed for GCN to perform well. 
To investigate this hypothesis, we use the randomization schemes described in Section~\ref{sec:randomization_strategy} to degrade
systematically the information content of the graph and/or the features in our data sets.
Fig.~\ref{fig:summary_randomization} presents the performance of the GCN as a function of the percent of randomization of the graph structure, the features, or both. 
As expected, the accuracy decreases for all data sets as the information contained in the graph, features or both is scrambled, yet with differences in the decay rate of each of the ingredients for the different examples.

Note that the chance-level performance of the ``Complete graph'' (mean field) limiting case is achieved only when \textit{both} graph and features are fully randomized, whereas the accuracy of the two other limiting cases (``No graph---MLP'', ``No features'') is reached around the half-point ($\sim 50\%$) of randomization of the graph or of the features, 
respectively. 
This indicates that using the scrambled information above a certain degree of randomization becomes more detrimental to the classification performance than simply ignoring it. 

\subsection{Relating GCN Performance and Subspace Alignment}
\begin{figure}[htbp!]
\centering
\includegraphics[width=0.75\textwidth]{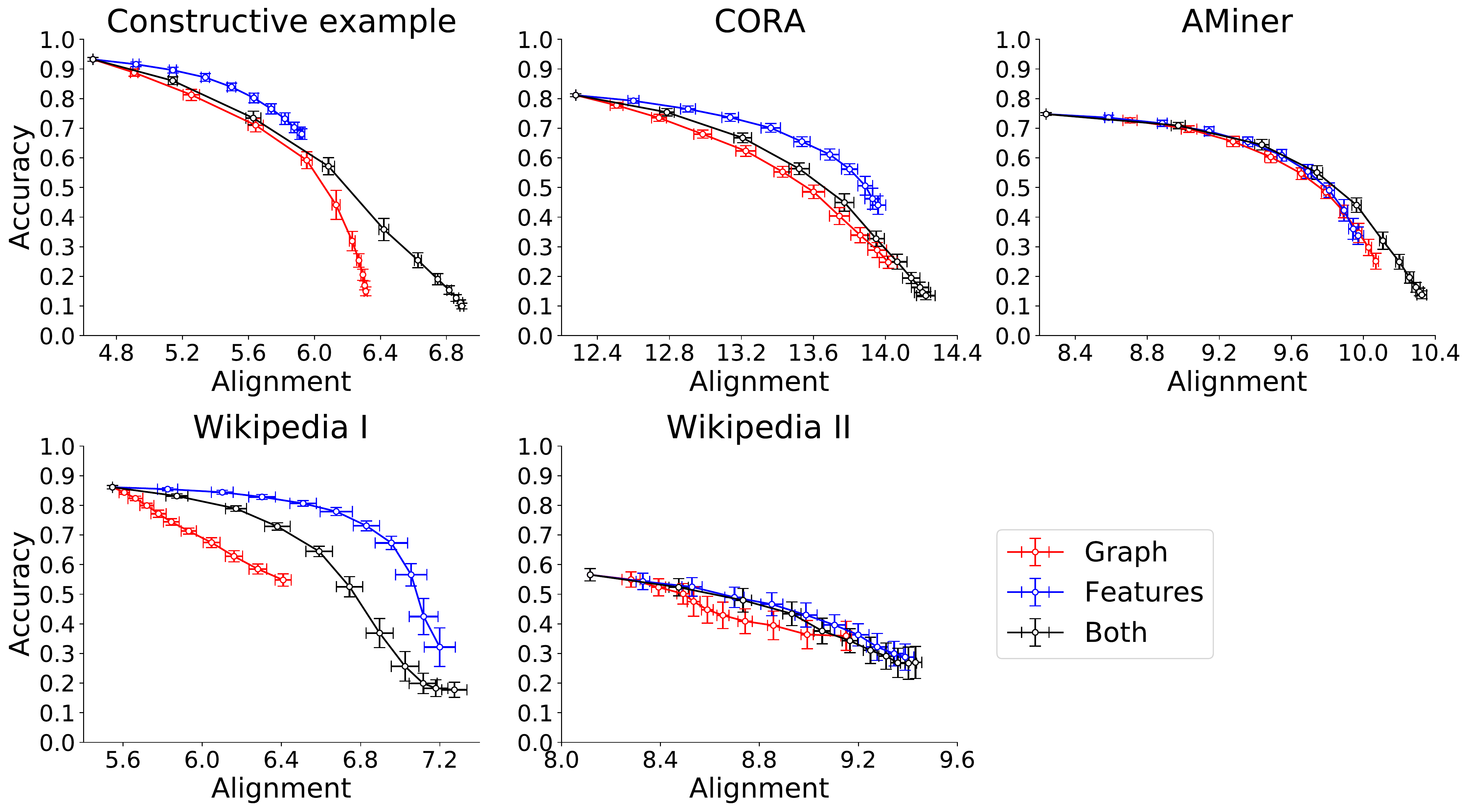}
\caption{\textbf{Classification performance versus the subspace alignment measure (SAM).} Each panel shows the accuracy of GCN versus the SAM~\eqref{eq:SAM} for all the runs presented in Fig.~\ref{fig:summary_randomization}.
Error bars are evaluated over $100$ randomizations.}
\label{fig:summary_Frobenius_norm}
\end{figure}

We tested whether the degradation of GCN performance is linked to the increased misalignment of features, graph and ground truth given by the SAM
\begin{equation}
\label{eq:SAM}
   \mathcal{S}^*(X,\widehat{A},Y) = \|D(X,\widehat{A},Y;k^{*}_{X}, k^{*}_{\widehat{A}}, k^*_Y) \|_{\text{F}} 
\end{equation}
which corresponds to~\eqref{eq:norm_matrix_distance} computed with 
the dimensions $(k^{*}_{X}, k^{*}_{\widehat{A}}, k^*_Y)$ obtained using~\eqref{eq:dimension_subspaces} (Table~\ref{table:k_X_k_A}, and see SI for the optimization scheme used). 
Fig.~\ref{fig:summary_Frobenius_norm} shows that the GCN accuracy is clearly (anti)correlated with the subspace alignment distance~\eqref{eq:SAM} in all our examples (mean correlation $= -0.92$).
As we randomize the graph and/or features, the subspace misalignment increases and the GCN performance decreases. 
In addition to the Chordal distance,~\cite{ye2016schubert} studies other subspace distances. While all the distances can be expressed in terms of the principal angles $\theta_j$, some rely on all the angles whereas others only use the maximum principal angle. We obtain similar results for distances that use all the principal angles (e.g., Chordal, Grassmann), but we find that extremal distances based on the maximum principal angle (e.g., the Projection distance) do not correlate as well with GCN performance. This highlights the importance of the information captured by all principal angles to quantify the alignment between subspaces. For results based on the Grassmann and Projection distances, see Appendix (Section~\RNum{3}) in the Supplementary Material.

\begin{table}[htbp!]
\centering
\resizebox{0.5\textwidth}{!}{
\begin{threeparttable}
\begin{tabular}{cccc}
\specialrule{.1em}{.05em}{.05em}
\textbf{Data sets} & $k^{*}_{X}$ & $~k^{*}_{\widehat{A}}$ & $~k^{*}_{Y}$
\\
\hline
Constructive example & 287 & 10 & 10\\
CORA & 1,291 & 190 & 7\\
AMiner & 500 & 57 & 7\\
Wikipedia~\RNum{1} & 68 & 1,699 & 5\\
Wikipedia~\RNum{2} & 100 & 1,125 & 5\\
\specialrule{.1em}{.05em}{.05em}
\end{tabular}
\end{threeparttable}}
\caption{\textbf{Dimensions of the three subspaces obtained according to~\eqref{eq:dimension_subspaces} for our data sets.}}
\label{table:k_X_k_A}
\end{table}

\section{DISCUSSION\label{sec:discussion}}

Our first set of experiments (see Table~\ref{table:results_gcn_and_limitCases}) reflects the varying amount of information that GCN can extract from features, graph and their combination, for the purpose of classification. 
For a classifier to perform well, it is necessary to find (possibly nonlinear) combinations of features that map differentially and distinctively onto the categories of the ground truth.
The larger the difference (or distance on the projected space) between the samples of each category, the easier it is to ``separate'' them, and the better the classifier.
In the MLP setting, for instance, the weights between layers ($W^\ell$) are trained to maximize this separation. As seen by the different accuracies in the ``No graph' 'column (Table~\ref{table:results_gcn_and_limitCases}), the features of each example contain variable amount of information that is mappable on its ground truth. 
A similar reasoning applies to classification based on graph information alone, but in this case, it is the eigenvectors of $\widehat{A}$ that need to be combined to produce distinguishing features between the categories in the ground truth
(e.g., if the graph substructures across scales~\cite{lambiotte2015random} do not map onto the separation lines of the ground truth categories, then the classification performance based on the graph will deteriorate). 
The accuracy in the ``No features'' column indicates that some of the graphs contain more congruent information with the ground truth than others.
Therefore, the ``No graph'' and ``No features'' limiting cases inform about the relative congruence of each type of information with respect to the ground truth. One can then conjecture that if the performance of the ``No features'' case is higher than the ``No graph'' case, GCN will yield better results than MLP.

In addition, our numerics show that although combining both sources of information generally leads to improved classification performance (``GCN original'' column in Table~\ref{table:results_gcn_and_limitCases}),
this is \textit{not} always necessarily the case.
Indeed, for the Wikipedia and Wikipedia~\RNum{2} examples, the classification performance of the MLP (``No graph''), which is agnostic to relationships between samples, is better than when the additional layer of relational information about the samples (i.e., the graph) is incorporated via the GCN architecture.
This suggests that, for improved GCN classification, the information contained in features and graph needs to be constructively aligned with the ground truth.
This phenomenon can be intuitively understood as follows.
In the absence of a graph (i.e., the MLP setting), the training of the layer weights is done independently over the samples, without assuming any relationship between them.  
In GCN, on the other hand, the role of the graph is to guide the training of the weights %
by averaging the features of a node with those of its graph neighbors. 
The underlying assumption is that the relationships represented by the graph should be consistent with the information of the features, i.e., the features of nodes that are graph neighbors are expected to be more similar than otherwise; hence the training process is biased towards convolving the diffusing information on the graph to extract improved feature descriptions for the classifier.
However, if feature similarities and graph neighborhoods (or more generally, graph communities~\cite{lambiotte2015random}) are not congruent, this graph-based averaging during the training is not beneficial. 

To explore this issue in a controlled fashion, 
our second set of experiments (Fig.~\ref{fig:summary_randomization}) studied the degradation of the classification performance induced by the systematic randomization of graph structure and/or features.
The erosion of information is not uniform across our examples, reflecting the relative salience of each of the components (features and graph) for classification.
Note that the GCN is able to leverage the information present in any of the two components, and is only degraded to chance-level performance when \textit{both} graph and features are fully randomized.
Interestingly, this fully randomized (chance-level) performance coincides with that of the ``Complete graph'' (or mean field) limiting case, where the classifier is trained on features averaged over all the samples, thus leading to a uniform representation that has zero discriminating power when it comes to category assignment.

These results suggest that a degree of constructive alignment between the matrices of features, graph and ground truth 
$(X, \widehat{A},Y)$ 
is necessary for GCN to operate successfully beyond standard classifiers. To capture this idea, we proposed a simple SAM~\eqref{eq:SAM} that uses the minimal principal angles to capture the consistency of pairwise projections between subspaces. Fig.~\ref{fig:summary_Frobenius_norm} shows that SAM correlates well with the classification performance and captures the monotonic dependence %
remarkably, given that SAM is a simple linear measure being applied to the outcome of a highly non-linear, optimized system. 
The results are consistent for other versions of GCN. In particular, in the Supplementary Material (Section~\RNum{2}) we show that the alignment measure correlates well with the performance of the recently proposed Simple Graph Convolution (SGC)~\cite{wu2019simplifying}.

The alignment measure can be used to evaluate the relative importance of features and graph for classification without explicitly running the GCN, by comparing the SAM under full randomization of features against the SAM under full randomization of the graph. If $\mathcal{S}^*(X_{100},\widehat{A},Y) > \mathcal{S}^*(X,\widehat{A}_{100},Y)$, the features play a more important role in GCN classification.
Conversely, if $\mathcal{S}^*(X_{100},\widehat{A},Y) < \mathcal{S}^*(X,\widehat{A}_{100},Y)$, the graph is more important in GCN classification. While we have focused here on node classification, it would be interesting in future work to extend our measure to other tasks such as graph classification, link prediction, and regression.

\section{CONCLUSION\label{sec:conclusion}}

Here, we have introduced SAM~\eqref{eq:SAM}, a measure that quantifies the consistency between the feature and graph ingredients of data sets, and we showed that it correlates well with the classification performance of GCNs. Our experiments show that a degree of alignment is needed for a GCN approach to be beneficial, and that using a GCN can actually be detrimental to the classification performance if the feature and graph subspaces associated with the data are not constructively aligned (e.g., Wikipedia and Wikipedia II). More generally, the SAM has potentially a wider range of applications in the quantification of data alignment including, among others: quantifying the alignment of different graphs associated with, or obtained from, particular data sets; evaluating the quality of classifications found using unsupervised methods; and aiding in choosing the classifier architecture most advantageous computationally for a particular data set. 

Our approach has a number of limitations that could be addressed in future work. First, it contains two parameters (i.e., the dimensions of the subspaces, $k_{X}^{*}$ and $k_{\widehat{A}}^{*}$) which need to be tuned through a computational search. Second, the alignment is not directly comparable across data sets since the subspace dimensions are adjusted for each data set. To facilitate comparisons across data sets, normalized versions of the alignment measure will be the object of future work. Third, the current measure is not suitable for very large data sets as the eigendecomposition of large matrices is computationally demanding. For very large data sets, approximations (e.g., using the Lanczos algorithm to explore only leading eigenvectors) might be necessary to optimize the subspace dimensions.

\section*{ACKNOWLEDGEMENT}
The work of Yifan Qian was supported by the China Scholarship Council Program under Grant 201706020176. The work of Paul Expert was supported in part by the National Institute for Health Research (NIHR) Imperial Biomedical Research Centre (BRC) under Grant NIHR-BRC-P68711 and in part by the Engineering and Physical Sciences Research Council (EPSRC) Centre for Mathematics of Precision Healthcare under Grant EP/N014529/1. The work of Mauricio Barahona was supported by the EPSRC Centre for Mathematics of Precision Healthcare under Grant EP/N014529/1.

\section*{BIOGRAPHY}
Yifan Qian received the B.Sc. degree in information and computing science and the M.Sc. degree in computer science from Beihang University, Beijing, China, in 2014 and 2017, respectively. He is currently pursuing the Ph.D. degree with the Queen Mary University of London, London, U.K. His research interest is broadly concerned with computational social science and combines theories and methods from network science, sociology, machine learning, and data science.

Paul Expert received the Ph.D. degree in physics from Imperial College London, London, U.K., in 2012. He was in Neuroimaging at King's College London, London, and Mathematics at Imperial College London. He is currently a Research Associate with the Global Digital Health Unit, Imperial College London, and a Visiting Associate Professor with the Tokyo Institute of Technology, Tokyo, Japan. His research interest is concerned with understanding the interaction between the structure and function of complex systems, with applications ranging from neuroscience to public health.

Tom Rieu received the M.Sc. degree in Engineering from the Engineering School CentraleSupélec Paris, France, and the M.Sc. degree in applied mathematics from the Imperial College London, London, U.K., both in 2017. He is currently a Data Scientist with Facebook, London, U.K. His academic research was focused on machine learning and particularly the application of deep models to networks. Currently, his work as part of the Facebook Product team is concerned with producing data science insights to drive improvements on the core and business products of Facebook’s online advertising platform and retargeting technology.

Pietro Panzarasa received the Ph.D. degree from Bocconi University Milan, Italy, in 2000.
He is a Professor of Networks and Innovation with the School of Business and Management, Queen Mary University of London, London, U.K. He became a Research Fellow with the University of Southampton, Southampton, U.K. He also held visiting positions at Columbia University, New York, NY, USA, and Carnegie Mellon University, Pittsburgh, PA, USA. He draws on network science, computational social science, and
big data analytics to study social capital and dynamics of social interaction in complex large-scale networks.

Mauricio Barahona received the Ph.D. degree from the Massachusetts Institute of Technology, Cambridge, MA, USA, in 1996. He is a Professor with the Department of Mathematics and the Director of the EPSRC Centre for Mathematics of Precision Healthcare, Imperial College London, London, U.K. He held Fellowships with Stanford University, Stanford, CA, USA, and the California Institute of Technology, Pasadena, CA. He is broadly interested in applied mathematics in engineering, physical, social, and biological systems using methods from graph theory, stochastic processes, dynamical systems, and machine learning.

\end{document}


\maketitle

This is the supplementary material to the article entitled ``Quantifying the alignment of graph and features in deep learning'', published in~\textit{IEEE Transactions on Neural Networks and Learning Systems}.

\section{Finding optimal dimensions}
A key element of the subspace alignment measure described in the main paper is to find lower dimensional representations of the graph, features and ground truth.

To determine the dimension of the representative subspaces, we propose the following heuristic:

\begin{align}
  (k_{X}^{*},k_{\widehat{A}}^{*})&=
  \underset{k_{X},k_{\widehat{A}}}{\max}
  \left(\|D(X_{100},\widehat{A}_{100},Y)\|_{\text{F}}-
  \|D(X,\widehat{A},Y)\|_{\text{F}}\right).
  \label{eq:dimension_subspaces} %
\end{align}

We choose $k_{Y}^{*}$ to be equal to the number of categories in the ground truth as they are non overlapping. Thus, $k_{X}^{*}$ and $k_{\widehat{A}}^{*}$ range from $k_{Y}^{*}$ to their maximum values, $C^{0}$, the dimension of the feature vectors, and $N$, the number of nodes in the graph, respectively.

\begin{figure}[H]
\centering
\begin{subfigure}[c]{.27\textwidth}
\includegraphics[width=1.0\textwidth]{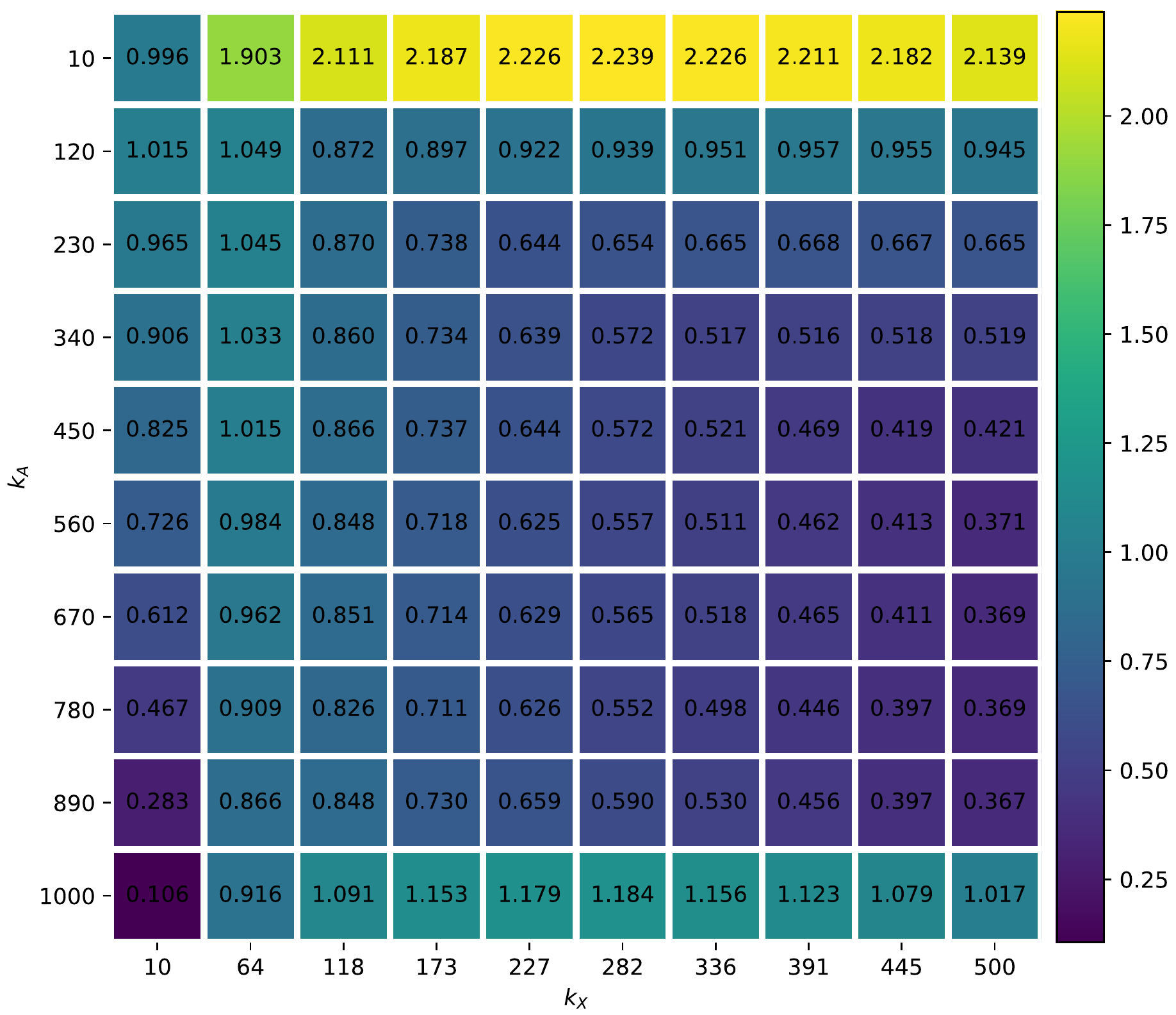}
\caption{Constructive example: round 1}
\end{subfigure}
~
\begin{subfigure}[c]{.27\textwidth}
\includegraphics[width=1.0\textwidth]{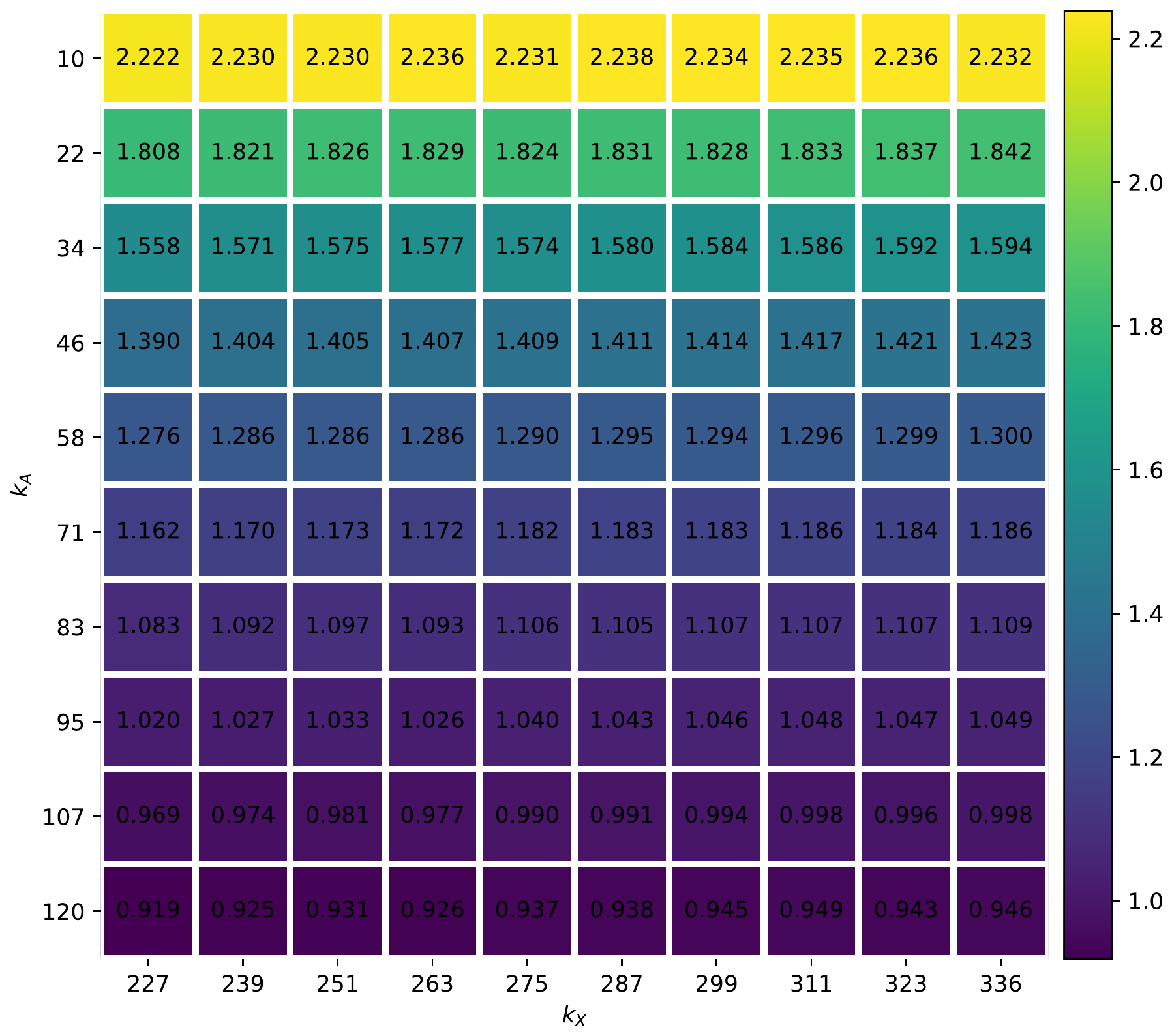}
\caption{Constructive example: round 2}
\end{subfigure}

\begin{subfigure}[c]{.27\textwidth}
\includegraphics[width=1.0\textwidth]{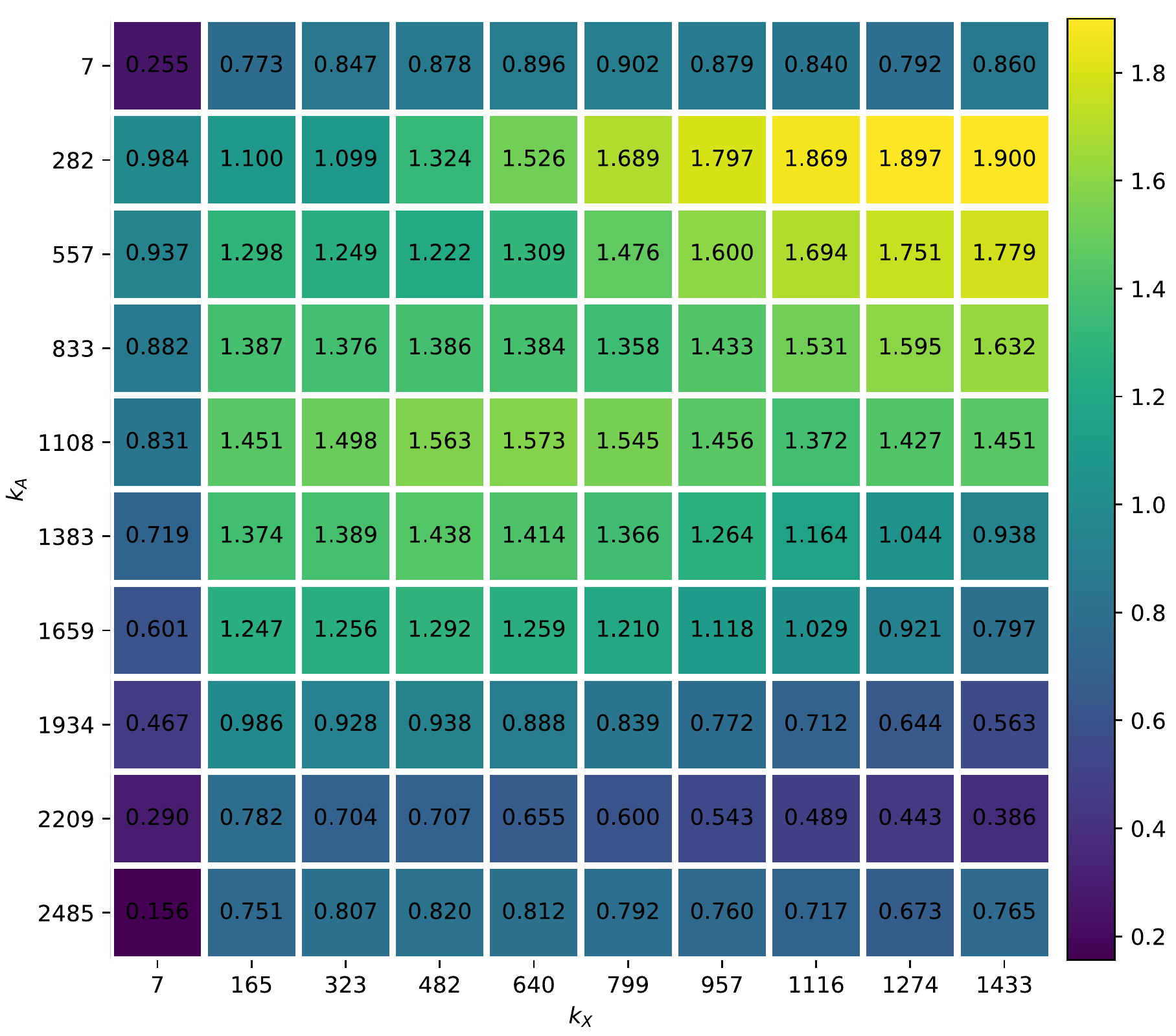}
\caption{CORA: round 1}
\label{CORA: round 1}
\end{subfigure}
~
\begin{subfigure}[c]{.27\textwidth}
\includegraphics[width=1.0\textwidth]{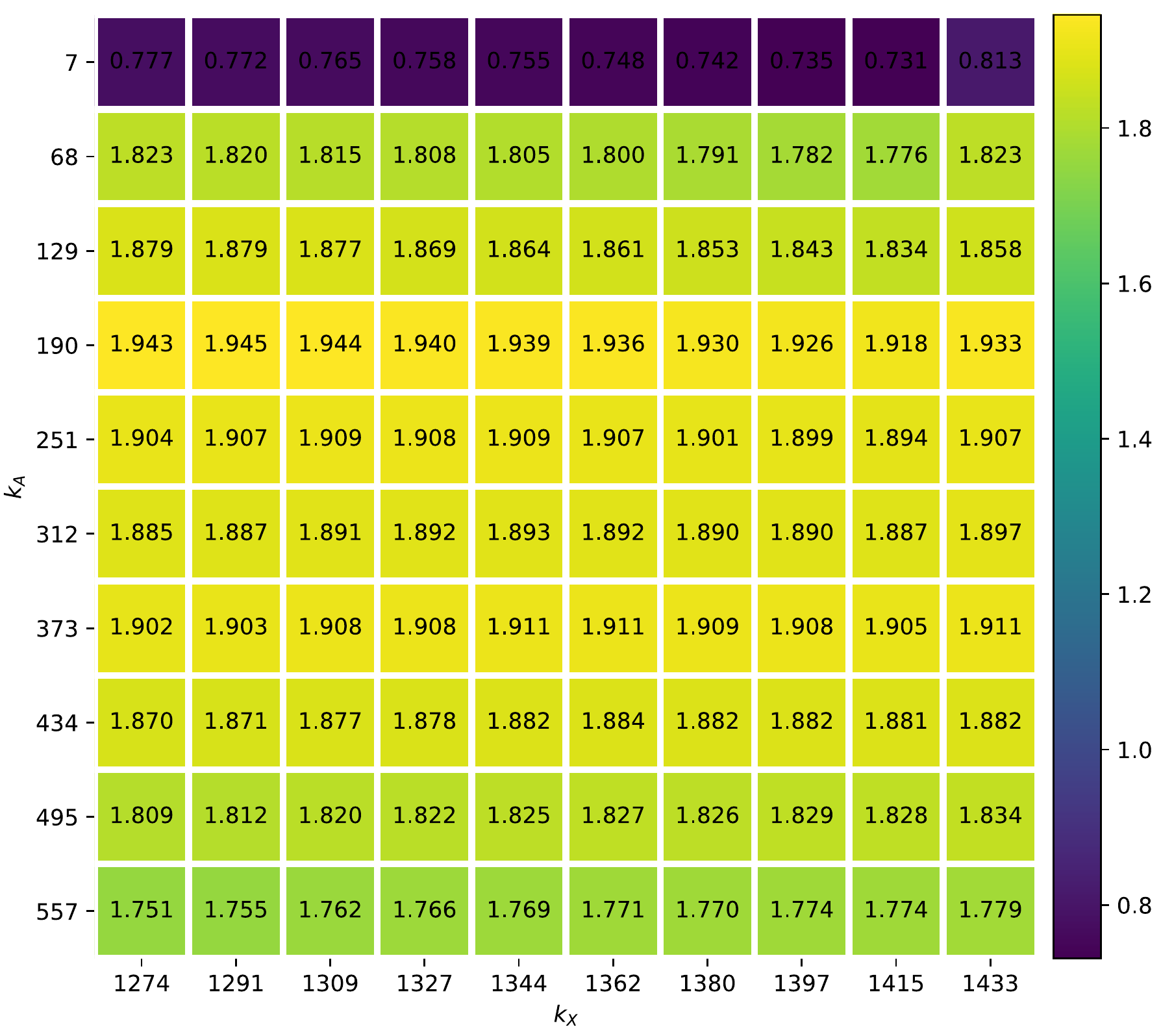}
\caption{CORA: round 2}
\label{CORA: round 2}
\end{subfigure}

\begin{subfigure}[c]{.27\textwidth}
\includegraphics[width=1.0\textwidth]{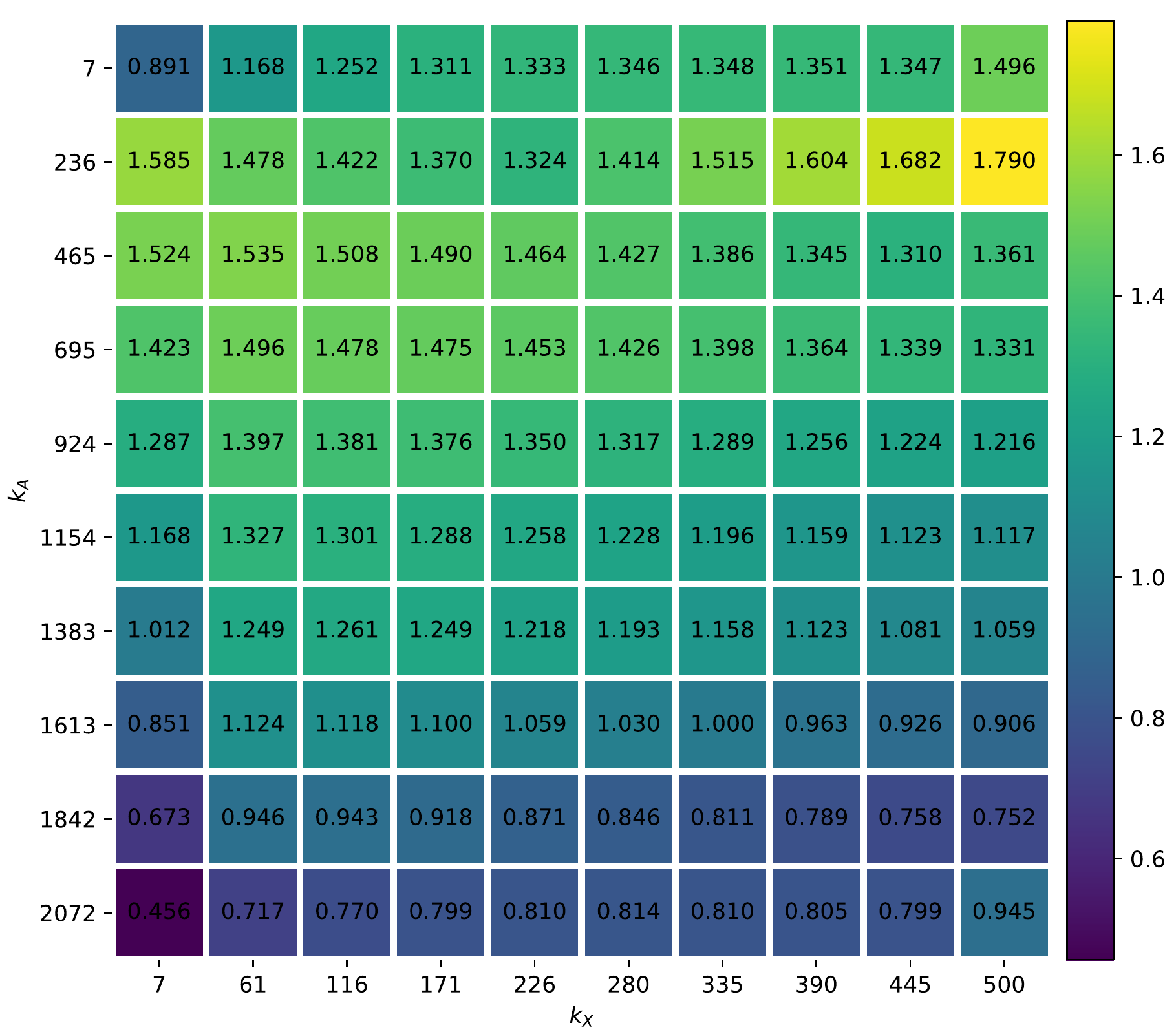}
\caption{AMiner: round 1}
\end{subfigure}
~
\begin{subfigure}[c]{.27\textwidth}
\includegraphics[width=1.0\textwidth]{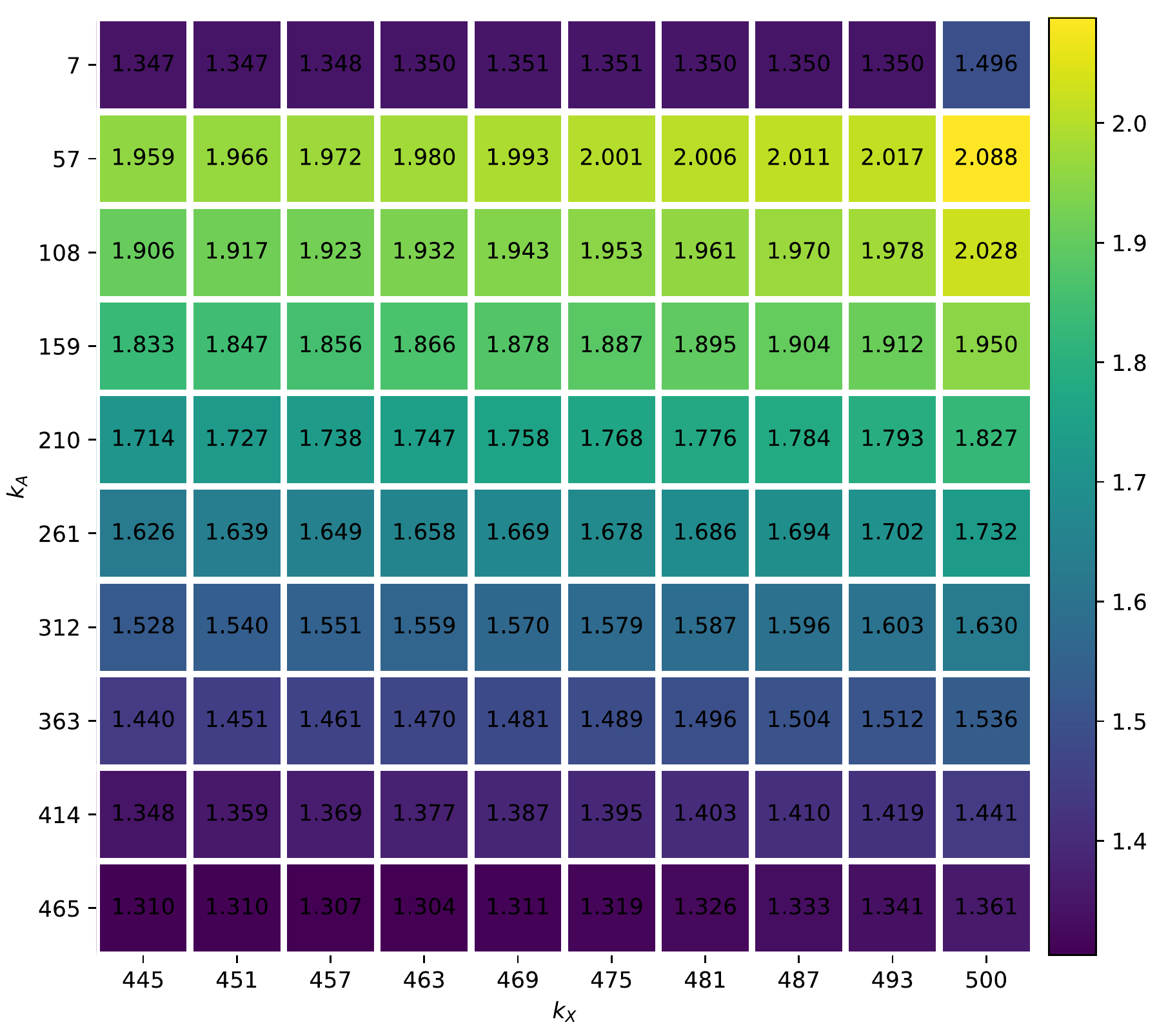}
\caption{AMiner: round 2}
\end{subfigure}

\begin{subfigure}[c]{.27\textwidth}
\includegraphics[width=1.0\textwidth]{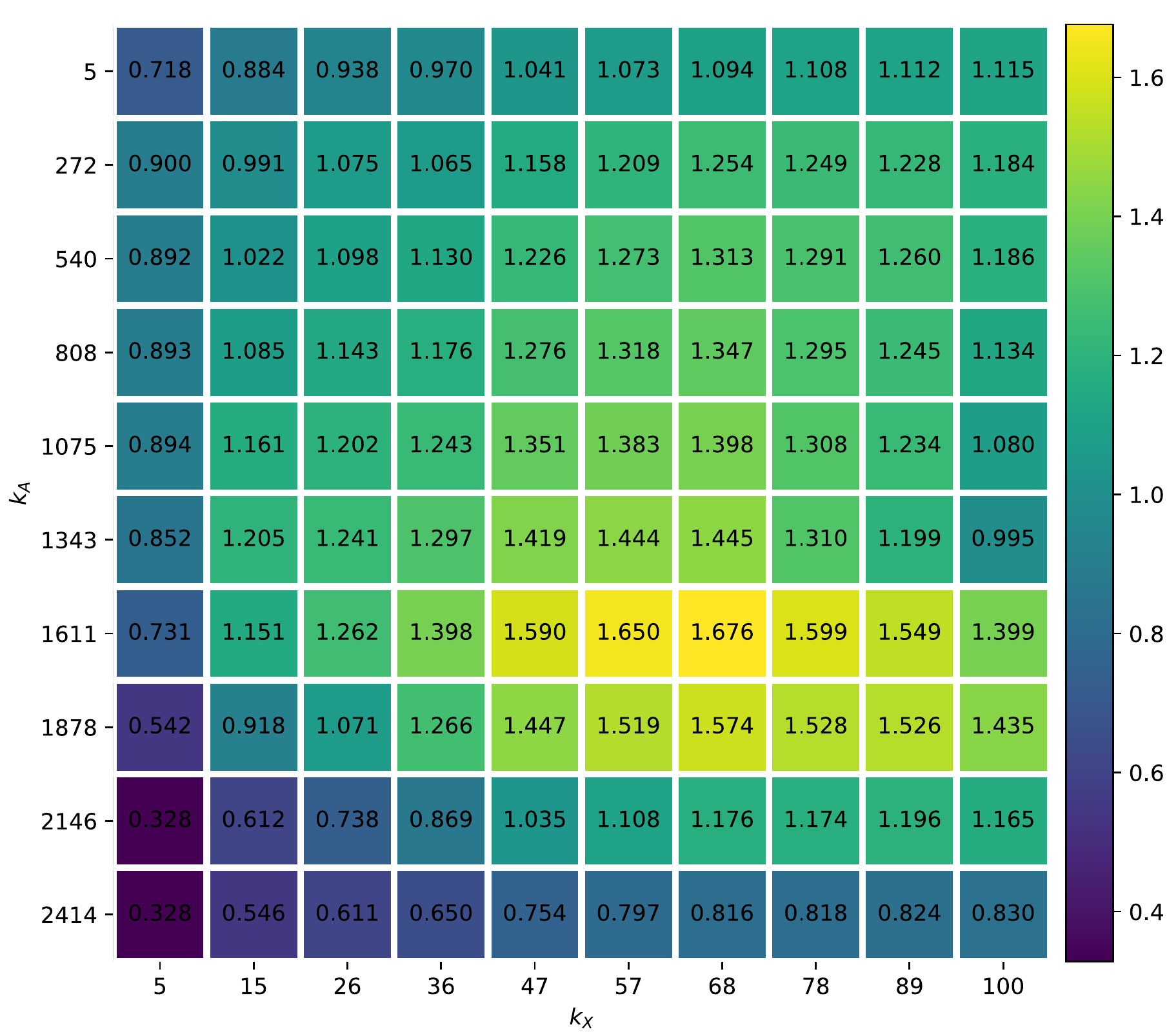}
\caption{Wikipedia~\RNum{1}: round 1}
\end{subfigure}
~
\begin{subfigure}[c]{.27\textwidth}
\includegraphics[width=1.0\textwidth]{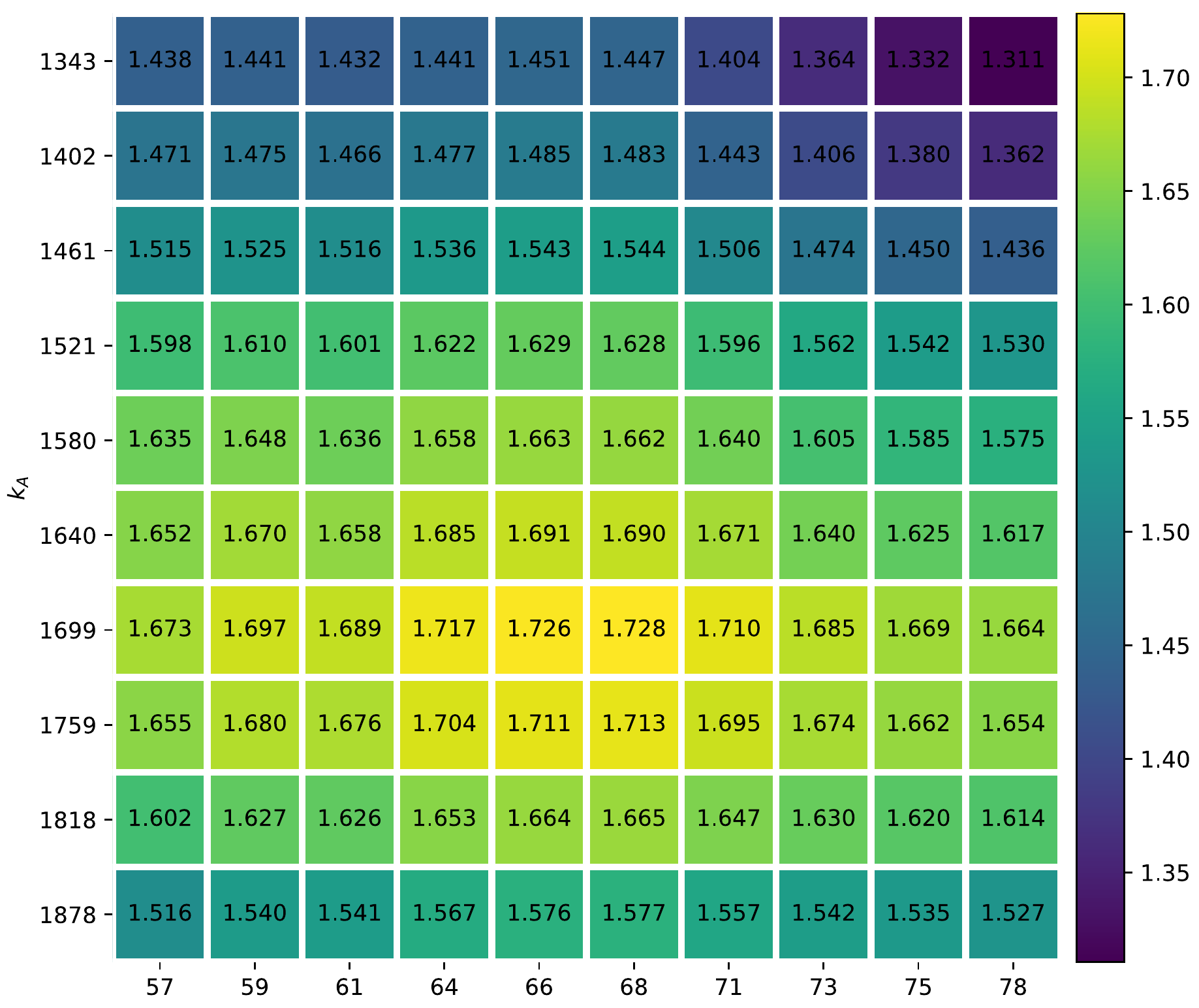}
\caption{Wikipedia~\RNum{1}: round 2}
\end{subfigure}

\begin{subfigure}[c]{.27\textwidth}
\includegraphics[width=1.0\textwidth]{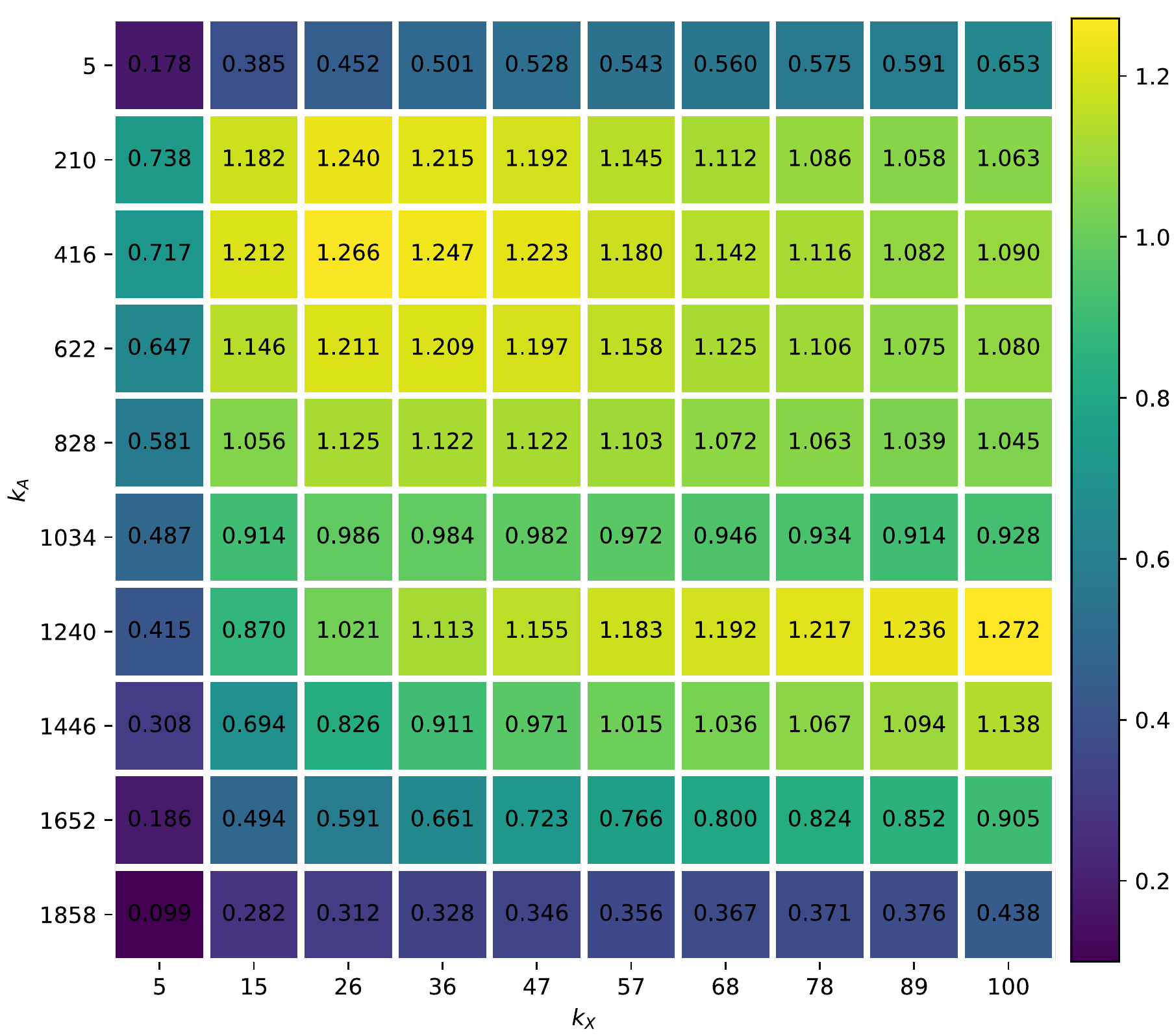}
\caption{Wikipedia~\RNum{2}: round 1}
\end{subfigure}
~
\begin{subfigure}[c]{.27\textwidth}
\includegraphics[width=1.0\textwidth]{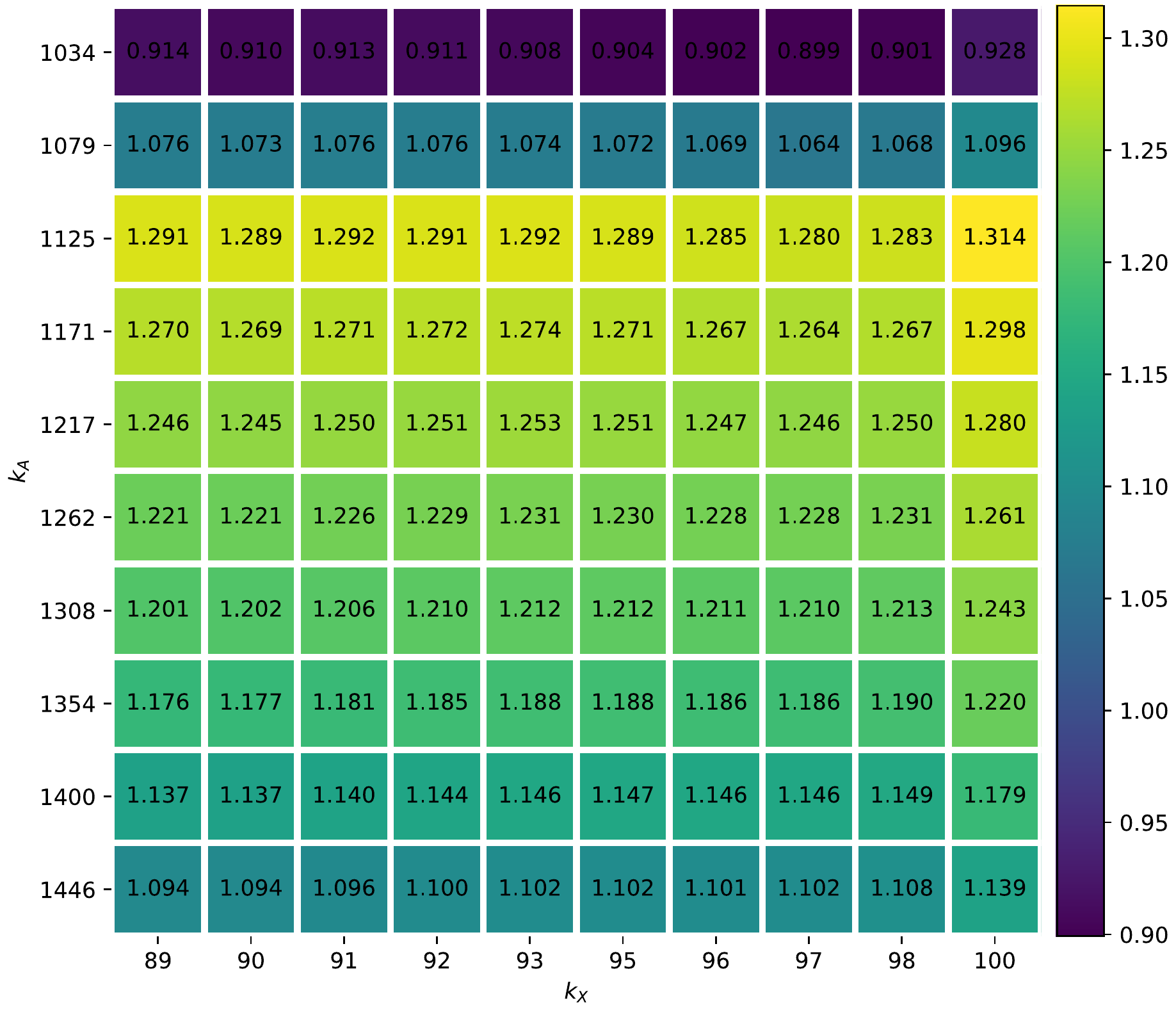}
\caption{Wikipedia~\RNum{2}: round 2}
\end{subfigure}
\caption{Summary of results on scanning subspaces.}
\label{fig:summary_scanning}
\end{figure}

To find the values for $k_{X}^{*}$ and $k_{\widehat{A}}^{*}$, we scan different possible combinations of $k_{X}$ and $k_{\widehat{A}}$. We applied two rounds of scanning. In the first scanning round, in the intervals of $k_{X}$ and $k_{\widehat{A}}$, we picked $10$ equally spaced values that contain the minimum and maximum possible values for $k_{X}$ and $k_{\widehat{A}}$. For example, in CORA, $k_{Y}^{*}$ equals $7$ because the number of categories in the ground truth is $7$. Thus $k_{X}$ ranges from $7$ to $1,433$.
At the end of the first round, the optimal values of $k_{X}^{*}$ and $k_{\widehat{A}}^{*}$ are $1,433$ and $282$, respectively (see Fig.~\ref{CORA: round 1}).

In the second scanning round, we applied a very similar process to the one just described. We set the scanning intervals of $k_{X}$ and $k_{\widehat{A}}$ as the neighbors of $k_{X}^{*}$ and $k_{\widehat{A}}^{*}$ found in the first round, respectively. For example, in CORA, for the second round, we set the intervals of $k_{X}$ and $k_{\widehat{A}}$ as $[1,274, 1,433]$ and $[7,557]$. Again, we split the new intervals with $10$ equally spaced values.
We have also shown the scanning results for other data sets in Figure~\ref{fig:summary_scanning}.

\section{Replicating experiments on a variant of Graph Convolutional Networks}
First, we would like to highlight that the alignment metric is independent of the architecture and only relies on the data. Therefore, we expect that the conclusion will be consistent with different variants of GCNs: the convolution operation in GCN (Kipf and Welling) can be seen as a neighborhood aggregation or message passing scheme. Many variants based on the Kipf and Welling version of GCN have been proposed, but they can ultimately be expressed as neighborhood aggregation or message passing schemes. For these different GCNs, we expect that our hypothesis that a certain degree of alignment between ingredients is needed for them to perform well would hold, since the working principles of variants of GCNs and the original version we consider are similar. To substantiate this claim, we have replicated our experiments using a recently proposed variant of GCN: Simple Graph Convolution (SGC) proposed by Wu et al. ``Simplifying graph convolutional networks." ICML (2019). SGC is a simplified version of the original GCN proposed by Kipf and Welling that removes nonlinearities and collapses weight matrices between consecutive layers. It has been shown that SGC can achieve competitive performance on node classification tasks and yields up to several orders of magnitude speedup.

We use the implementation provided by Pytorch Geometric\footnote{https://github.com/rusty1s/pytorch\_geometric/blob/master/examples/sgc.py}, which is a popular geometric deep learning extension library for PyTorch. Our results on SGC are shown below in Fig.~\ref{fig:summary_randomization_sgc_SI} and Fig.~\ref{fig:summary_Frobenius_norm_sgc_SI}. The figure suggests that results based on SGC are consistent with those produced using GCN (Kipf and Welling).
\begin{figure}[htbp!]
    \centering
    \includegraphics[width=0.75\textwidth]{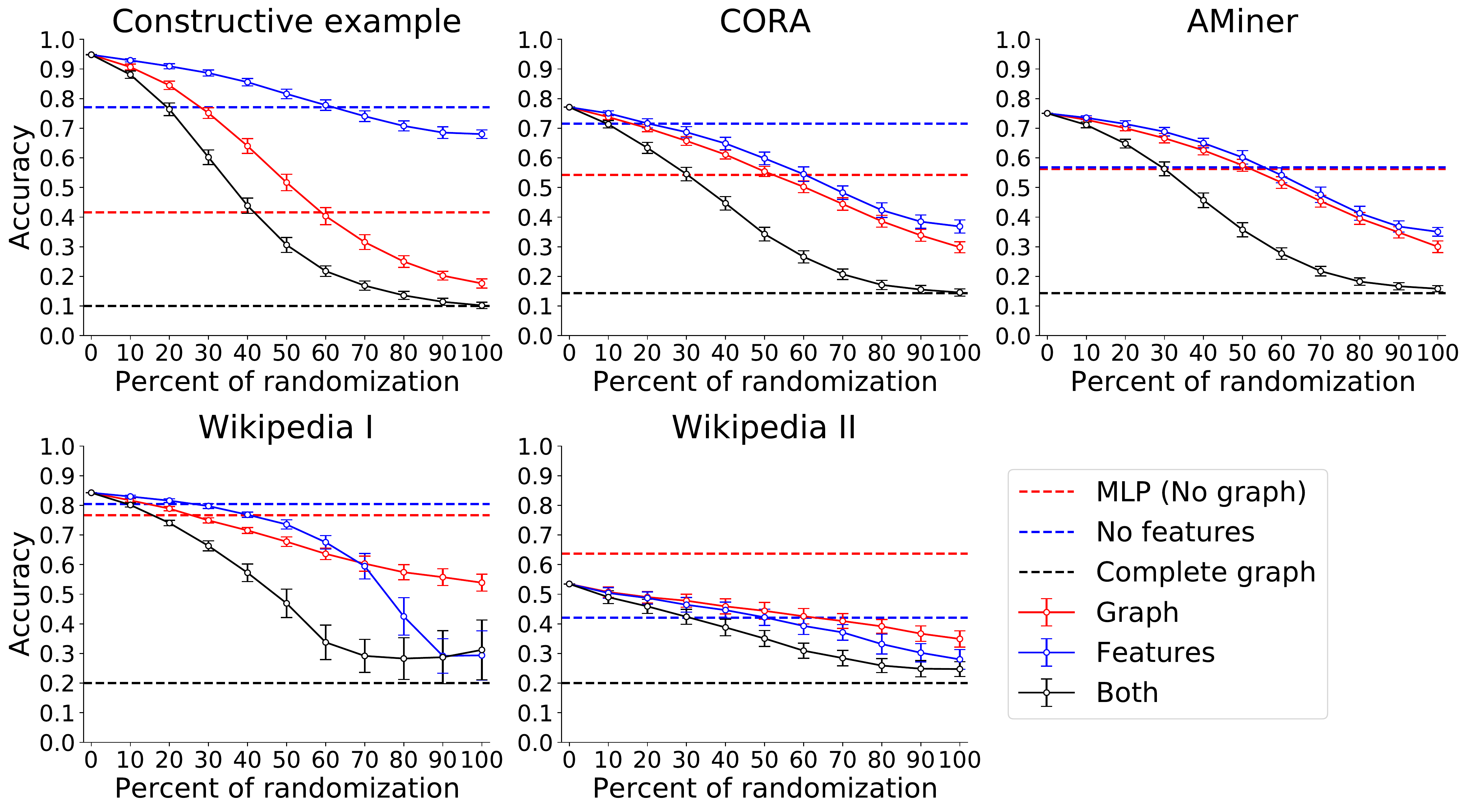}
    \caption{\textbf{Degradation of the classification performance as a function of randomization with SGC.} Each panel shows the degradation of the classification accuracy as a function of the randomization of graph, features and both, for a different data set. Error bars are evaluated over $100$ realizations:
    for zero percent randomization, we report $100$ runs with random seeds; for the rest, we report $1$ run with random seed for $100$ random realizations. The horizontal lines correspond to the limiting cases.
    }
    \label{fig:summary_randomization_sgc_SI}
\end{figure}
\begin{figure}[htbp!]
    \centering
    \includegraphics[width=0.75\textwidth]{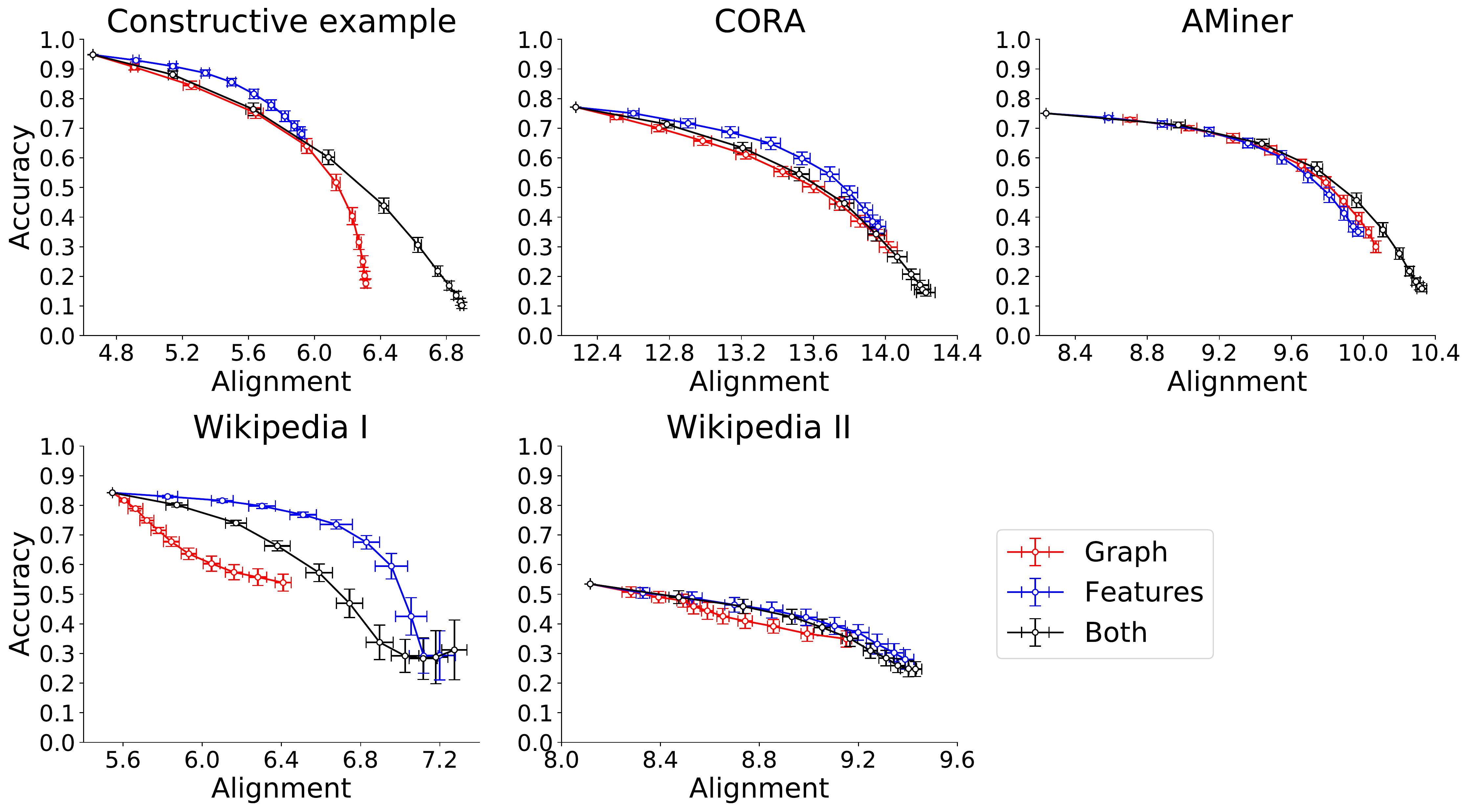}
    \caption{\textbf{Classification performance versus the subspace alignment measure (SAM) with SGC.} Each panel shows the accuracy of SGC versus the SAM for all the runs presented in Fig.~\ref{fig:summary_randomization_sgc_SI}. Error bars are evaluated over $100$ randomizations.}
    \label{fig:summary_Frobenius_norm_sgc_SI}
\end{figure}

\section{Choices of distance measures}

Within the distances discussed by Ref. (Ye, Ke, and Lim, Lek-Heng. ``Schubert varieties and distances between subspaces of different dimensions." SIAM Journal on Matrix Analysis and Applications 37.3 (2016): 1176-1197.), there are two `families':

\begin{enumerate}
\item average distances that use \textit{all} the principal angles, e.g., the Chordal distance, $\left(\sum_{j=1}^{\alpha} \sin ^{2} \theta_{j}\right)^{1 / 2}$, and the Grassmann distance, $\left(\sum_{j=1}^{\alpha} \theta^{2}_{j}\right)^{1 / 2}$.
\item extremal distances that use only the maximum principal angle between two subspaces, e.g., the Projection distance, $\sin \theta_{\alpha}$where $\theta_\alpha$ is the maximum angle.
\end{enumerate}
Our numerics show that average distances (the first family) display similar performance, as they leverage information from all the principal angles. Hence these measures produce similar performance to the Chordal distance. 
To show this, we have replicated our experiments using the Grassmann distance (see Fig.~\ref{fig:summary_Frobenius_norm_grassmann_SI} below). The results are consistent with those produced with Chordal distance. \\~\\           
On the other hand, we expect that extremal distances (the second family) will have less expressive power to capture the alignment between subspaces, since they use solely the maximum principal angle and do not consider the information contained in the other principal angles. To demonstrate this point, we replicated our experiments with the Projection distance (see Fig.~\ref{fig:summary_Frobenius_norm_projection_SI} below). Our results show that the Projection distance is indeed less effective than the Chordal distance in representing the alignment between subspaces.

\begin{figure}[htbp!]
      \centering
      \includegraphics[width=0.75\textwidth]{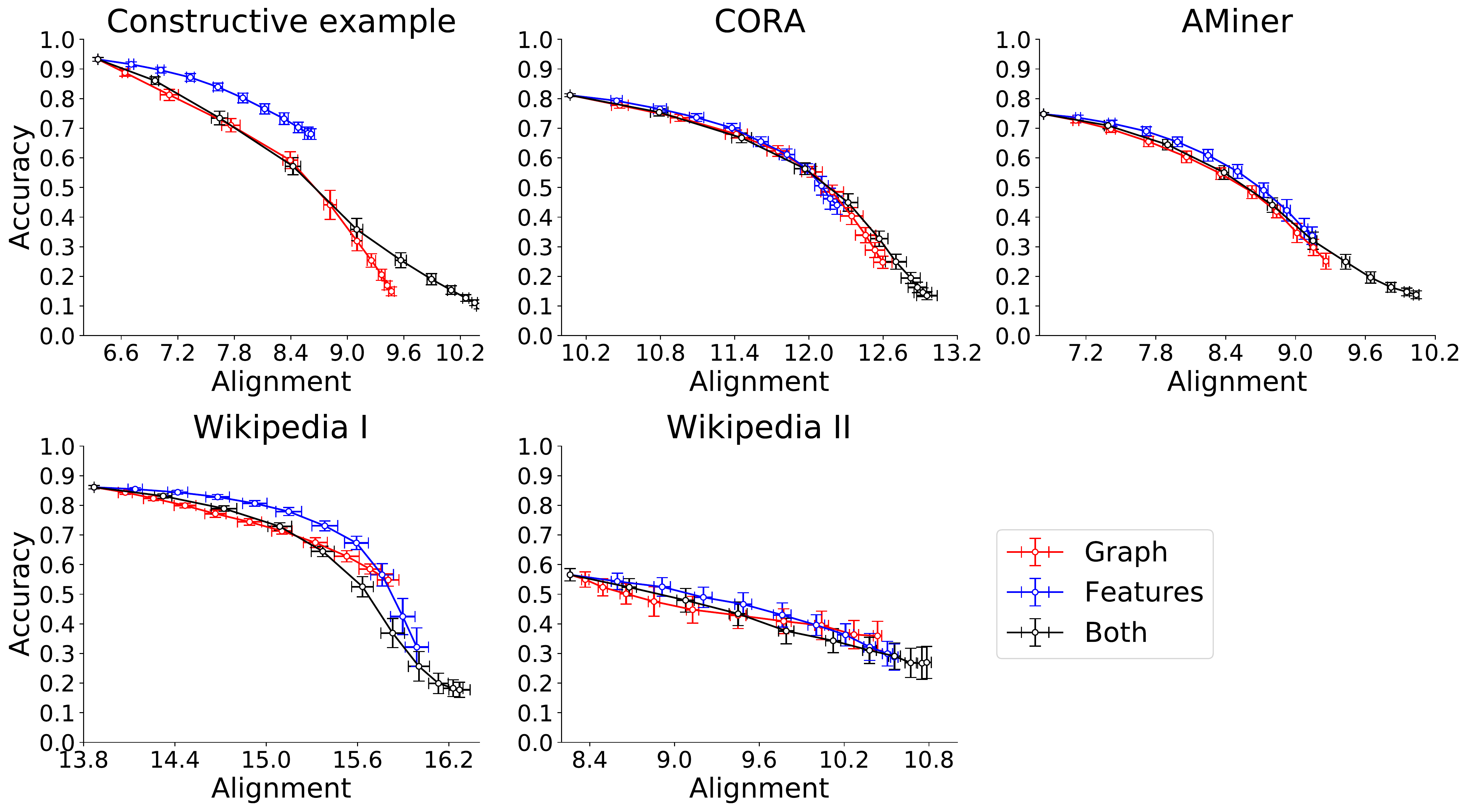}
      \caption{\textbf{Classification performance versus the subspace alignment measure (SAM) with Grassmann distance.} Each panel shows the accuracy of GCN versus the SAM. Error bars are evaluated over $100$ randomizations.}
      \label{fig:summary_Frobenius_norm_grassmann_SI}
\end{figure}

\begin{figure}[htbp!]
      \centering
      \includegraphics[width=0.75\textwidth]{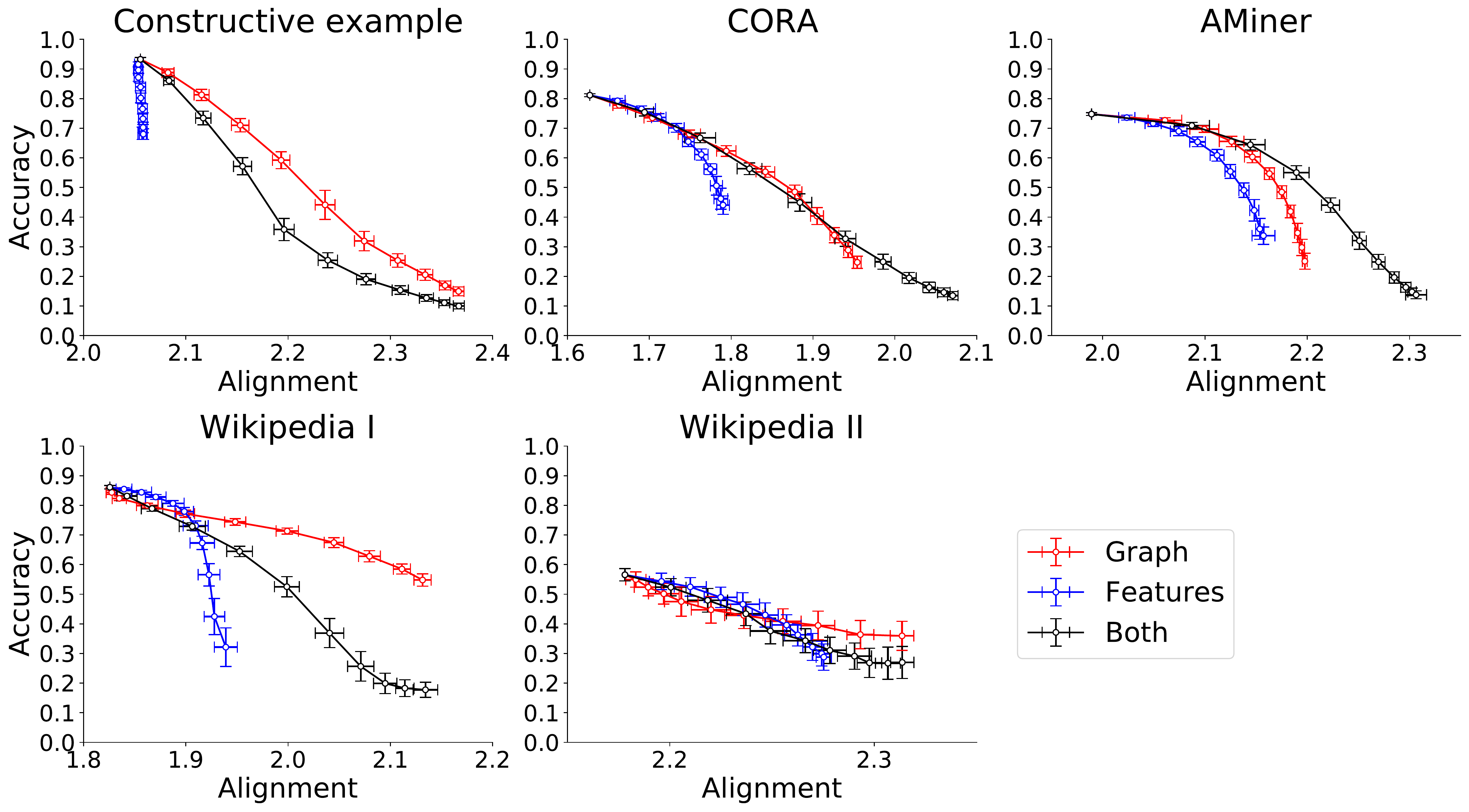}
      \caption{\textbf{Classification performance versus the subspace alignment measure (SAM) with Projection distance.} Each panel shows the accuracy of GCN versus the SAM. Error bars are evaluated over $100$ randomizations.}
      \label{fig:summary_Frobenius_norm_projection_SI}
\end{figure}